\documentclass[10pt,twocolumn,letterpaper]{article}

\usepackage{cvpr}            
\usepackage{makecell}

\usepackage{algorithm}
\usepackage{algorithmic}
\usepackage{booktabs,siunitx,subcaption,xcolor}
\usepackage[table]{xcolor} 
\usepackage[normalem]{ulem}
\usepackage{booktabs,multirow}
\usepackage{tcolorbox}
\definecolor{cvprblue}{rgb}{0.9,0.1,0.74}
\usepackage[pagebackref,breaklinks,colorlinks,allcolors=cvprblue]{hyperref}

\newcommand{\webpage}{\url{https://snap-research.github.io/vptt}~}
\title{Visual Personalization Turing Test}

\author{Rameen Abdal\\
Snap Research \\
\and
James Burgess\\
Stanford University\\
\and
Sergey Tulyakov\\
Snap Research\\
\and
Kuan-Chieh Jackson Wang\\
Snap Research \\
\and
{\tt\small \webpage}
}

\begin{document}
\maketitle
\begin{abstract}
We introduce the Visual Personalization Turing Test (VPTT), a new paradigm for evaluating contextual visual personalization based on perceptual indistinguishability, rather than identity replication. A model passes the VPTT if its output (image, video, 3D asset, etc.) is indistinguishable to a human or calibrated VLM judge from content a given person might plausibly create or share. To operationalize VPTT, we present the VPTT Framework, integrating a 10k-persona benchmark (VPTT-Bench), a visual retrieval-augmented generator (VPRAG), and the VPTT Score, a text-only metric calibrated against human and VLM judgments. We show high correlation across human, VLM, and VPTT evaluations, validating the VPTT Score as a reliable perceptual proxy. Experiments demonstrate that VPRAG achieves the best alignment–originality balance, offering a scalable and privacy-safe foundation for personalized generative AI.
\end{abstract}
\begin{figure}[t] 
    \centering \includegraphics[width=0.7\linewidth]
    {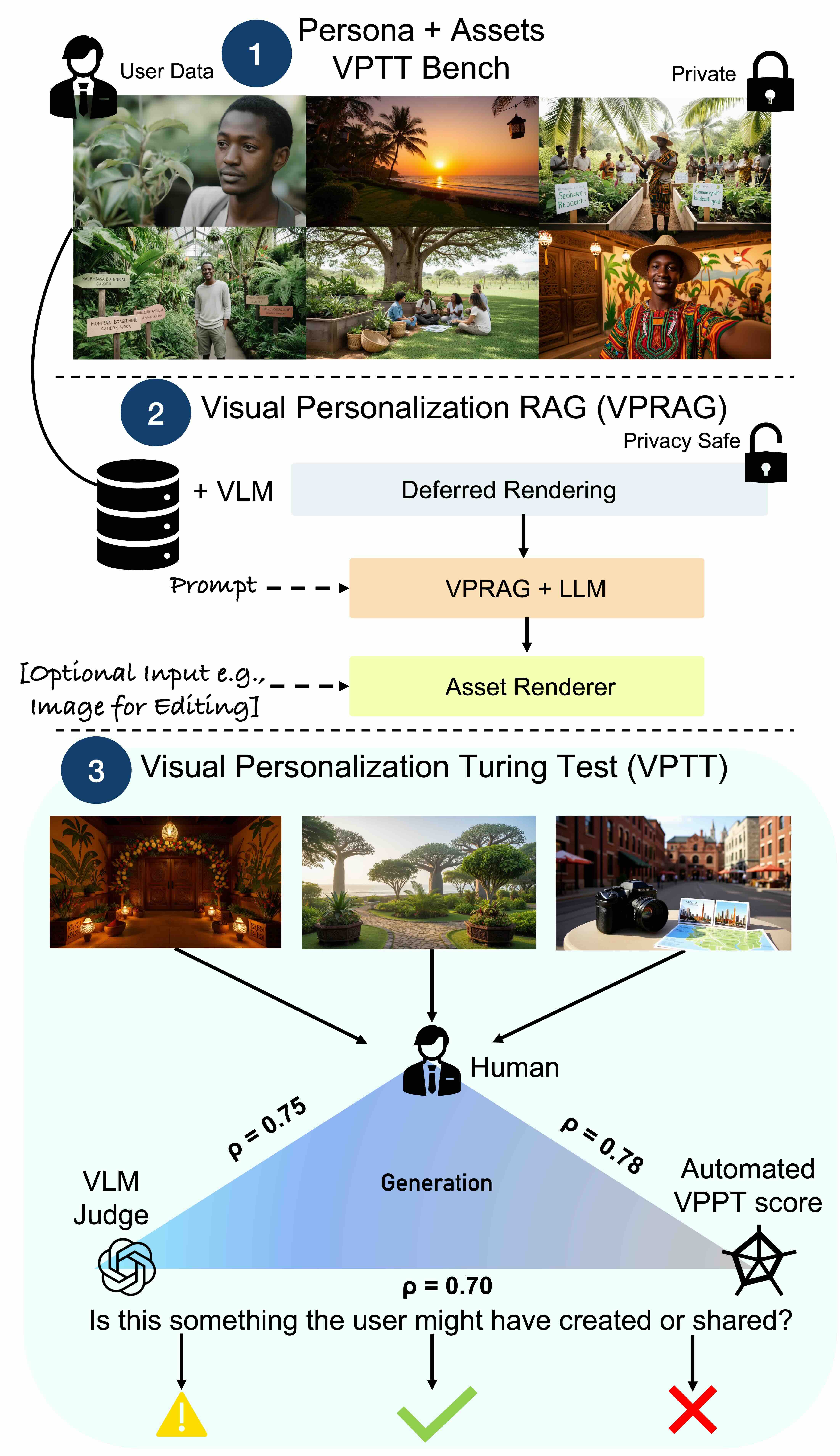}
    \caption{\textbf{Visual Personalization Turing Test.} We present the Visual Personalization Turing Test (VPTT), a new paradigm for contextual personalization at scale. A model passes the VPTT if its output is indistinguishable to a human or a calibrated VLM judge from what a given person might plausibly create or share. As one way to address this challenge, we introduce \textit{VPTT Framework} consisting of privacy-safe benchmark \textit{VPTT-Bench} for evaluating personalized generation and editing, and Visual Personalization RAG (\textit{VPRAG}) that retrieves persona-aligned visual cues and converts them into personalized image generations or edits. To close the loop, we propose an automated $\mathrm{VPTT_{score}}$ that achieves strong Spearman rank correlation ($\rho$) with humans and VLM Judges, establishing it as a cheap, reliable proxy for human perception of personalization.}
    \label{fig:vptt}
\vspace{-0.7cm}
\end{figure}    
\vspace{-0.8cm}
\section{Introduction} \label{sec:intro}

\begin{figure*}[t] 
    \centering \includegraphics[width=1.0\linewidth]{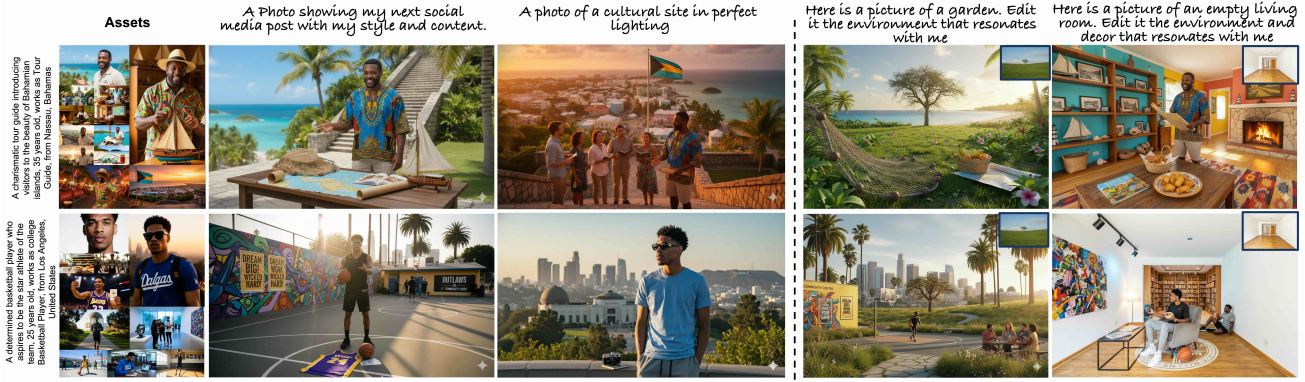}
    \vspace{-0.7cm}
    \caption{\textbf{Contextual Image Generation and Editing using VPTT-Bench.} Each row shows a distinct user profile: assets and style cues (left), personalized generations (social post, cultural site), and edits (garden, living room) guided by the same persona identity. All images are generated synthetically via our Visual Personalization RAG (VPRAG) by text, which retrieves persona-aligned cues. To show cross model personalization here the assets are generated by QWEN-image-model~\cite{yang2025qwen3technicalreport} and generations and edits by Nano-Banana~\cite{nano_banana} conditioned only on the first image. More results in are in Supplementary materials.}
    \label{fig:main}
    \vspace{-0.5cm}
\end{figure*}

Personalization in visual generation has so far focused on \emph{identity replication}~\cite{ruiz2023dreambooth,textual_inversion,custom-diffusion,subject-diffusion,wang2023high,gal2023designing, abdal2019image2stylegan, chen2024videoalchemy,abdal2025dynamic, ruiz2023hyperdreambooth, abdal2025zeroshotdynamicconceptpersonalization}, optimizing models to reproduce a subject across scenes. While effective at preserving appearance, these pipelines are computationally expensive ~\cite{ruiz2023dreambooth,abdal2025dynamic,textual_inversion} and miss the  broader vision of personalization: \emph{how individuals perceive, stylize, and share their world}. To instantiate this idea, personalization should capture the aesthetic preferences~\cite{ViPer, wang2025omnistylefilteringhighquality, openai2024gpt4technicalreport}, cultural context, and visual familiarity that constitute a person’s unique visual language. Yet, no benchmark exists to measure whether a model’s output truly \emph{feels like it could have been created by a particular person or a creator}. This gap is increasingly important beyond research. Industry is actively trying to bridge the gap between GenAI and user-created content to make generative AI \emph{monetizable, trustworthy, and personally resonant}~\cite{sora_model,veo2}. This challenge becomes even more pressing as powerful foundation models in image domain, such as Qwen~\cite{yang2025qwen3technicalreport},  NanoBanana~\cite{nano_banana} and GPT-Image-1~\cite{gpt_image}, already achieve near-photorealistic quality. As models master realism, the frontier of innovation shifts to what is personally relevant to the user~\cite{sora_model2}.

To address this gap, we introduce the Visual Personalization Turing Test (VPTT) (Figure.~\ref{fig:vptt}): a new paradigm for evaluating generative models. A model passes the VPTT if its output (image, video, 3D asset etc.) is \emph{indistinguishable to a human or a calibrated VLM judge from that a given person might plausibly create or share}. This reframes the goal from rote memorization of appearance to the far more challenging task of simulating a personal perspective.

Solving the VPTT presents three fundamental challenges. First, it requires a benchmark with thousands of diverse, culturally, and stylistically rich user profiles, yet real-world user data is inaccessible due to privacy concerns, fundamentally limiting academic research. Second, it demands a new technical approach beyond the fine-tuning one that can interpret a user's complex, multi-faceted style from their history and apply it to new generations in a scalable, efficient manner. Third, it requires a robust evaluation protocol to test VPTT at large scale. 

We introduce the \textit{VPTT framework}, designed to address these challenges at scale. To overcome the data barrier, we construct \textit{VPTT-Bench}, the first large-scale benchmark of about 10,000 synthetic personas, whose visual worlds (30 assets - images for the scope of this paper) are represented entirely in text as ``deferred renderings," (structured, attribute-rich intermediates like lighting , materials, environment, actions, forground, background, appearance etc. that defer visual realization, analogous to G-buffers~\cite{10.1145/378456.378468} in graphics) enabling privacy-safe research at scale. Additionally, we render about 1000 synthetic personas to create a rich visual library. As a possible solution to personalization at scale, we propose a novel visual personalization retrieval-augmented generation (\textit{VPRAG}) system. Instead of costly retraining, our method conditions generation on a persona's existing assets through hierarchical semantic retrieval with an optional learnable feedback and composes a personalized prompt enriched with their unique stylistic elements. 

Our evaluation framework for image generation and editing is two fold. We first introduce $\mathrm{VPTT_{score}}$ as a automatic proxy for VPTT. We conduct a visual-level evaluation through VPTT, validated by human study and extended with calibrated VLM judges. This helps us establish strong correlations among all three evaluators text-level ($\mathrm{VPTT_{score}}$), VLM, and human, confirming that the $\mathrm{VPTT_{score}}$ is a reliable, perceptually grounded proxy for visual judgment. After establishing this, we perform a large-scale deferred rendering analysis (about 120,000 evaluations) using the $\mathrm{VPTT_{score}}$. Our results show that VPRAG’s structured design achieves the best trade-off between output alignment and novelty, addressing a key limitation of black-box baselines. Our contributions are:

\begin{itemize}
    \item A new task formulation, the Visual Personalization Turing Test (VPTT), redefines success in visual personalization as achieving human indistinguishable authenticity.

    \item VPTT Framework, the first scalable, privacy-safe benchmark for contextual personalization, featuring 10,000 rich personas with 1,000 visually rendered agents.

    \item A novel Visual Personalization Retrieval-Augmented Generation (VPRAG) system, a structured, zero-shot engine for personalization offering a possible scalable solution.

    \item A rigorous new evaluation framework featuring the VPTT score validated against human and VLM judges, proving it is a reliable proxy for perceptual alignment.

    \item A comprehensive analysis on our benchmark using a mix of closed- and open-source models with varying computational budgets, demonstrating that VPRAG offers a better trade-off between performance and efficiency.
\end{itemize}

\section{Related Work}
\label{sec:related_work}

\begin{figure*}[t] 
    \centering \includegraphics[width=1.0\textwidth]{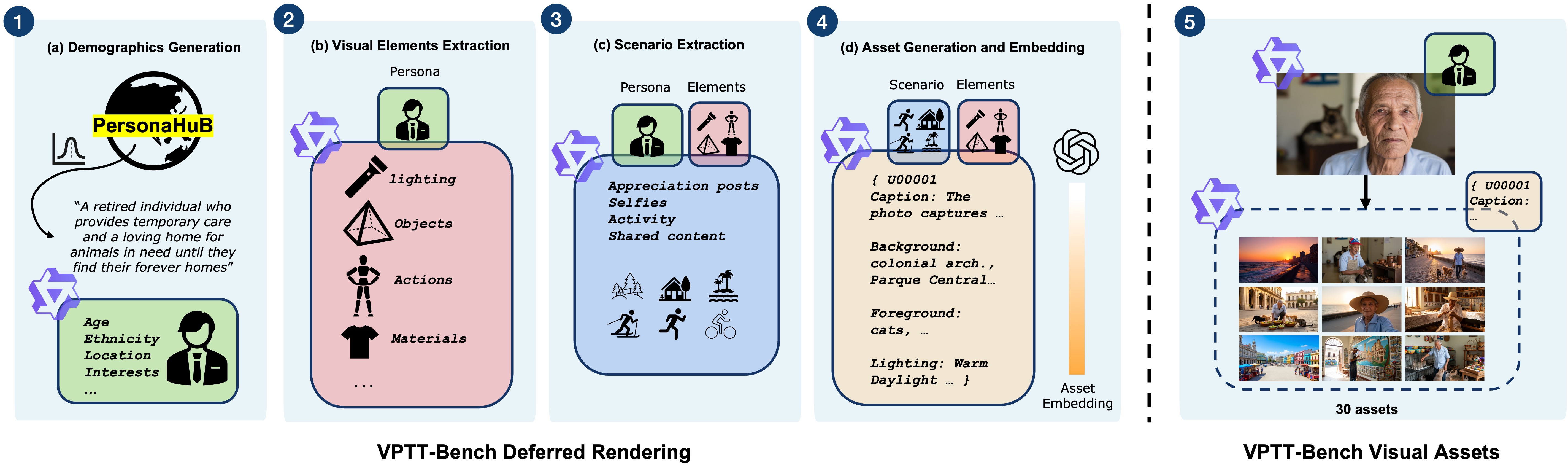}
    \vspace{-0.8cm}
    \caption{\textbf{VPTT-Bench Data Generation Pipeline.}
Overview of the deferred rendering pipeline used to construct VPTT-Bench.
(1) Personas are sampled from PersonaHub~\cite{ge2025scalingsyntheticdatacreation} with demographics.
(2–3) Visual and scenario elements (lighting, actions, materials etc.) are extracted.
(4) These cues are composed into structured captions and embedded via an LLM.
(5) Generating 30 corresponding visual assets per persona, forming privacy-safe, semantically grounded data for evaluating contextual personalization.}
\label{fig:data_gen}
\vspace{-0.5cm}
\end{figure*}

\begin{figure}[t] 
    \centering \includegraphics[width=0.5\textwidth]{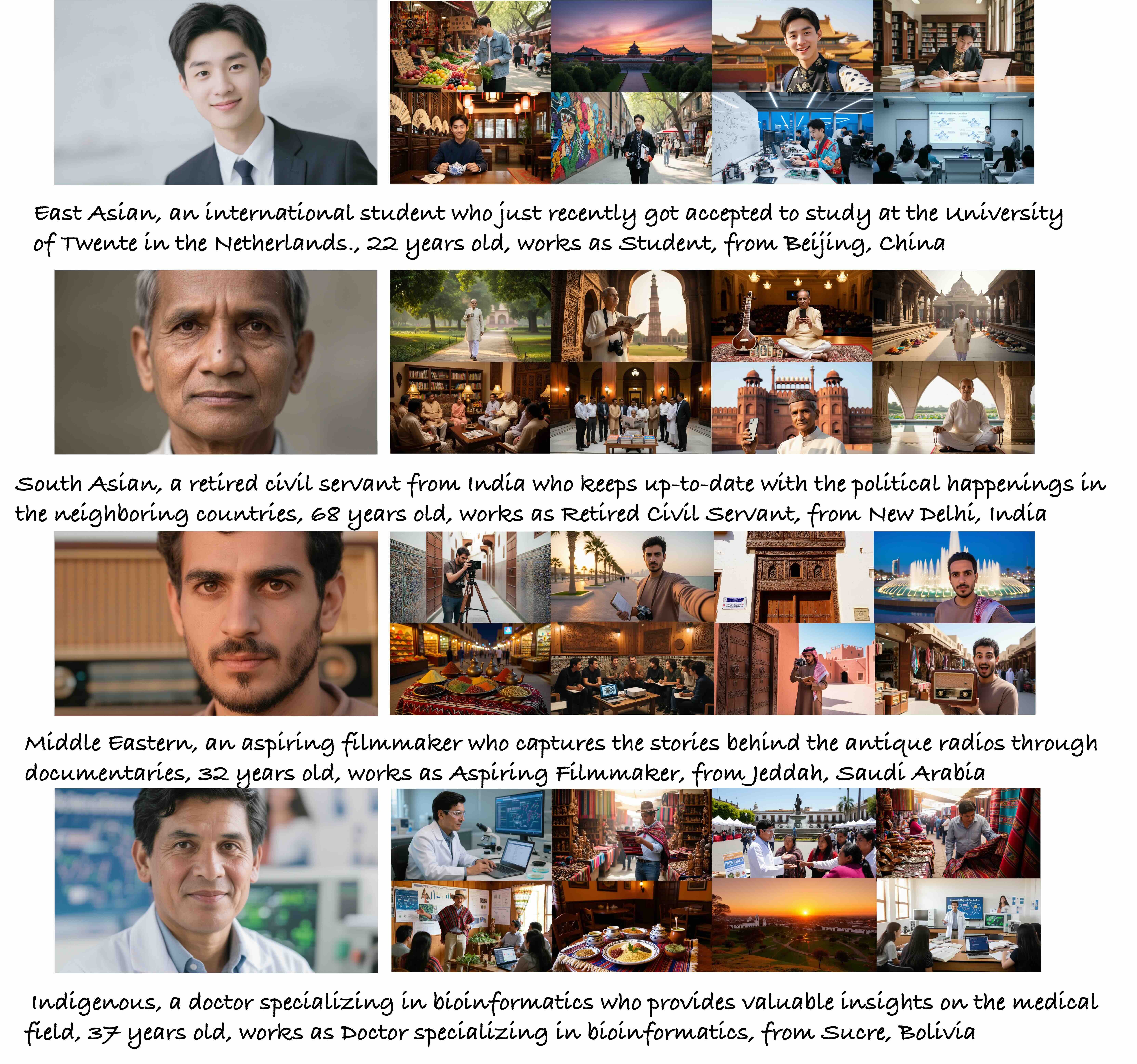}
    \vspace{-0.5cm}
    \caption{ \textbf{Example Personas from VPTT-Bench.}
Each row shows a synthetic persona sampled from PersonaHub~\cite{ge2025scalingsyntheticdatacreation} (only short descriptions) with its corresponding visual assets generated via VPTT-Bench generation pipeline.
Personas span diverse regions, professions, and age groups, illustrating the demographic and contextual diversity of VPTT-Bench.}
    \label{fig:personas}
    \vspace{-0.7cm}
\end{figure}

\subsection{Personalization in Visual Generative Models.}

Personalization in generative models has traditionally focused on identity replication~\cite{ruiz2023dreambooth,textual_inversion,custom-diffusion,abdal2019image2stylegan,chen2024videoalchemy,abdal2025dynamic,gal2024lcm,gal2022image,abdal2025zeroshotdynamicconceptpersonalization}. Seminal methods like DreamBooth~\cite{ruiz2023dreambooth} and LoRA adaptations~\cite{simo} excel at fine-tuning models to reproduce a specific subject across different scenes. However, these approaches are not scalable and primarily address appearance fidelity rather than the user's broader visual signature~\cite{ruiz2023dreambooth,textual_inversion,custom-diffusion,subject-diffusion,fastcomposer, tuning-encoder,shi2023instantbooth,wang2023high,gal2023designing, abdal2019image2stylegan, wang2024moa, gal2024lcm, gal2022image,ruiz2023hyperdreambooth}. More recent works aim for tuning-free personalization. IP-Adapter and related works in the image and video domains~\cite{chen2024videoalchemy,abdal2025zeroshotdynamicconceptpersonalization,fastcomposer,tuning-encoder,shi2023instantbooth,gal2024lcm, ye2023ip-adapter} use reference images to condition generation, achieving strong results in transferring style or appearance~\cite{hertz2023StyleAligned, frenkel2024implicit,gao2025styleshot, hu2024instructimagenimagegenerationmultimodal} but often requiring careful selection of reference images and suffer from the absence of a larger visual context~\cite{gao2025styleshot}. Methods like InstantBooth~\cite{shi2023instantboothpersonalizedtexttoimagegeneration} represent another direction in test-time personalization without fine-tuning but again focuses on personalizing the appearance of the subject. Among the methods that consider the context,  DrUM~\cite{kim2025drawmindpersonalizedgeneration} proposes learning a vector based on prompt history and injecting it via a trained adapter network, offering a modular approach but still involving per user adapter training.  A very recent work ImageGem~\cite{guo2025imagegeminthewildgenerativeimage}, collects in-the-wild interactions for generative model personalization, highlighting the community's growing interest in this area, though primarily focused on LoRAs collected over users generated content. Our work, orthogonal to these works, focuses on deriving and applying preferences, cultural context, visual familiarity and personal elements implicitly derived from a user's asset history, without requiring explicit reference images or per-user training of adapters. 

\subsection{ Visual Preference Personalization} 

Aligning generative models with user preferences is a critical challenge. Many recent efforts draw inspiration from Reinforcement Learning from Human Feedback (RLHF)~\cite{ouyang2022traininglanguagemodelsfollow} used in LLMs~\cite{comanici2025gemini25pushingfrontier,yang2025qwen3technicalreport}. An early work, ImageReward~\cite{xu2023imagerewardlearningevaluatinghuman} trained a reward model on human comparisons to score prompt-image alignment, enabling fine-tuning via Reward Feedback Learning (ReFL). Diffusion-DPO~\cite{wallace2023diffusionmodelalignmentusing} applied Direct Preference Optimization to fine-tune Stable Diffusion XL~\cite{podell2023sdxlimprovinglatentdiffusion} on large-scale human judgments~\cite{wu2023human} from datasets like Pick-a-Pic~\cite{kirstain2023pickapicopendatasetuser}, improving general appeal and alignment. While powerful, these methods typically optimize for aggregate preferences rather than individual context. On the other hand, approaches targeting individual preferences are emerging~\cite{nabati2025preferenceadaptivesequentialtexttoimage}. ViPer~\cite{ViPer} learns preferences by having an MLLM~\cite{openai2024gpt4technicalreport} analyze user comments on images, extracting structured attributes to guide generation. PPD~\cite{dang2025personalizedpreferencefinetuningdiffusion} trains a single model conditioned on user embeddings derived from few-shot pairwise preferences. POET~\cite{han2025poetsupportingpromptingcreativity} focuses on identifying image homogeneity using ``prompt inversion" and personalizing diversification based on interactive user feedback. Concurrent work, such as Instant Preference Alignment~\cite{li2025instantpreferencealignmenttexttoimage}, also uses MLLMs to extract preferences from a reference image for tuning-free guidance. Our work differs by focusing on extracting and applying alignment implicitly from a user's historical creative output (simulated via \textit{VPTT-Bench} derived from real-world grounded  PersonaHuB~\cite{ge2025scalingsyntheticdatacreation}) rather than relying on explicit feedback, pairwise comparisons, or single reference images. We introduce the VPTT as a holistic measure of visual context alignment beyond simple preference scores.

\begin{figure*}[t] 
    \centering \includegraphics[width=1.0\textwidth]{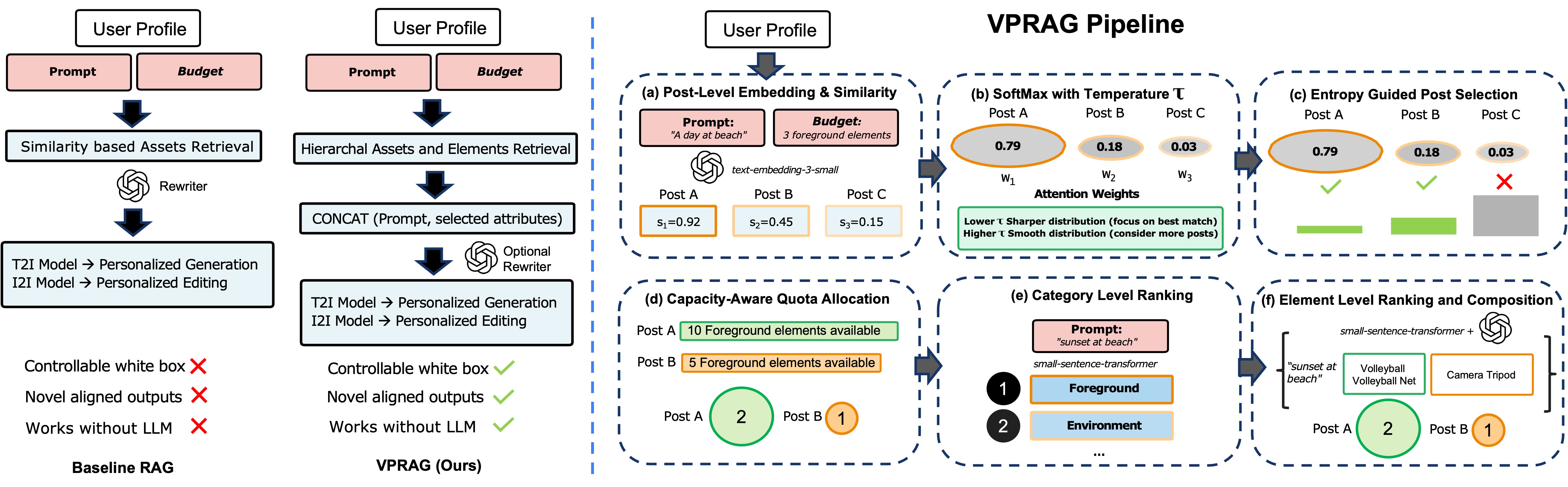}
    \vspace{-0.7cm}
    \caption{\textbf{VPRAG Pipeline Overview.}
Comparison between the baseline retrieval-augmented generation (BRAG) and our proposed Visual Personalization RAG (VPRAG).
Unlike baseline BRAG, VPRAG introduces controllable and interpretable retrieval through:
(a) post-level embedding and similarity scoring, (b) temperature-controlled attention, (c) entropy-guided post selection, (d) capacity-aware quota allocation, (e) category-level ranking, and (f) element-level composition.
This multi-stage design yields a white-box, LLM-optional retrieval framework producing visually and semantically aligned personalized generations and edits.}
    \label{fig:pipeline}
    \vspace{-0.4cm}
\end{figure*}

\subsection{RAG in Computer Vision} 

Retrieval-Augmented Generation (RAG)~\cite{gao2024retrievalaugmentedgenerationlargelanguage}, initially prominent in NLP, is increasingly being explored in computer vision~\cite{rag_web, shalevarkushin2025imageragdynamicimageretrieval}. Very recent works like RealRAG~\cite{lyu2025realragretrievalaugmentedrealisticimage} and FineRAG~\cite{yuan-etal-2025-finerag} focused on retrieving external visual knowledge (e.g., real images of objects) to improve content completion of generated images and using RAG for VQA. Comprehensive repositories like Awesome-RAG-Vision~\cite{rag_web} are mapping the growing landscape, covering applications in visual understanding, generation, and embodied AI. Within generation, RAPO~\cite{gao2025devilpromptsretrievalaugmentedprompt} uses RAG specifically for text-to-video prompt optimization, retrieving terms from a graph built on training data to align user prompts with the model's expected input format. Tailored Visions~\cite{chen2024tailoredvisionsenhancingtexttoimage} pioneered using RAG on a user's own prompt history for personalized text-to-image prompt rewriting, using an LLM to synthesize past styles into new prompts. OmniStyle~\cite{wang2025omnistylefilteringhighquality}, while focused on style transfer, utilizes a large curated dataset and filtering for high-quality supervised training. Our VRAG system builds upon the personalized RAG concept but distinguishes itself through: (1) operating on our structured, synthetic \textit{VPTT-Bench} benchmark, enabling privacy-safe research; and (2) employing a principled, more transparent retrieval and composition architecture for fine-grained control, rather than relying solely on a black-box LLM operating on raw prompt history.

\section{Visual Personalization Turing Test}
\label{sec:method}

Our goal is to model and evaluate \emph{contextual visual personalization} the ability of a generative model to produce content that a human (or VLM) would perceive
as consistent with a given persona’s visual context.
We formalize this as the Visual Personalization Turing Test (VPTT) and introduce
VPTT Framework, a unified framework that enables
systematic study of this problem at scale. VPTT Framework consists of four interacting components:
(1) a large-scale simulated persona benchmark (Sec.~\ref{sec:persona_plus});
(2) a retrieval-augmented generation engine (Sec.~\ref{sec:vrag});
(3) an optional learnable feedback loop (Sec.~\ref{sec:feedback});
and (4) a differentiable proxy metric, VPTT score (Sec.~\ref{sec:pgp}).
Together they form a closed cycle of simulation → generation → judgment → optimization.
\paragraph{Problem Definition.} Given a persona $\mathcal{P}=\{d,E,C\}$ demographics $d$, a structured element library $E$,
and a caption memory $C$ and a query $p$, the model must generate a personalized prompt $p'$
whose resulting image $\mathcal{G}(p')$ maximizes perceived alignment with $\mathcal{P}$:
\begin{equation}
\begin{aligned}
\mathcal{J}(p';\mathcal{P}) &=
\lambda_1\,\text{Align}(p',\mathcal{P})
+ \lambda_2\,\text{Fidelity}(p',C)  \\
&\quad +\, \lambda_3\,\text{Novelty}(p',C),
\qquad \sum_i \lambda_i = 1.
\end{aligned}
\label{eq:vptt}
\end{equation}
This surrogate defines the latent VPTT objective: an ideal system achieves high alignment,
high fidelity, and high novelty simultaneously, an intractable trade-off for current models. We expect this trade-off to improve with better personalized models and for the scope of this work propose a method that approximates this objective efficiently without retraining.

\subsection{VPTT-Bench: Scalable Simulation Substrate}
\label{sec:persona_plus}

Human personalization datasets are private and unscalable.
We therefore construct \textbf{VPTT-Bench} (Figure.~\ref{fig:data_gen} and Figure.~\ref{fig:personas}), a synthetic benchmark of $10{,}000$ agents,
each represented by a tuple $\mathcal{P}_i=\{d_i,E_i,C_i\}$.
Personas are generated using \texttt{Qwen2.5-72B-Instruct}~\cite{yang2025qwen3technicalreport}: \\
\textbf{Demographic Generation:} starting from public textual seeds (PersonaHUB~\cite{ge2025scalingsyntheticdatacreation}),
    we sample culturally diverse backstories $d_i$. This ensures cross-domain coverage, avoiding dataset bias. \\
\textbf{Visual Elements Extraction:} we sample and cluster atomic visual terms
    (e.g., clothing, lighting, pose) into structured vocabularies $E_i$ conditioned on $d_i$ ensuring the visual elements are consistent with the persona. \\
\textbf{Scenario and Assets Extraction:} conditioned on $\{d_i,E_i\}$, we first generate short scenarios of the assets and finally generate $30$ captions $C_i$ describing element rich posts with the scenario story arc. The captions are embedded using \texttt{text-embedding-3-small}~\cite{openai2024gpt4technicalreport}.

We further render a 1,000-persona subset into image galleries (30 images per persona), each anchored by a canonical portrait followed by caption-guided edits.
This hybrid text–image corpus provides both semantic control and visual diversity: the text-only component enables dense, scalable supervision without privacy constraints, while the paired visual assets allow controlled studies across different resource budgets, from lightweight text-only personalization to more expensive multimodal (text + image) setups. For real profiles, the reverse of this process is performed to get the structured data.

\subsection{Visual Personalization Retrieval-Augmented Generation (VPRAG)}
\label{sec:vrag}

To personalize content without model retraining,
we propose \textbf{VPRAG} (see Figure.~\ref{fig:pipeline}), a retrieval-augmented generation framework that conditions
prompt rewriting on a persona’s structured memory.
Given a query $p$ and profile $\mathcal{P}=\{d,E,C\}$,
VPRAG retrieves semantically relevant posts and elements, allocates retrieval quotas,
and composes a new prompt $p'$ that aligns with the persona’s context. Unlike other methods~\cite{dblora, ruiz2023dreambooth} that require minutes to hours per user, VPRAG operates entirely at inference time, adding only a few hundred milliseconds of retrieval and composition overhead.
\vspace{-0.4cm}
\paragraph{Hierarchical Retrieval.}
Captions $C$ encode holistic semantic intent (high-level concepts),
while elements $E$ capture atomic style (low-level cues). We therefore perform a hierarchical two-level retrieval for robustness.

\textit{\textbf{Post-level retrieval.}}
Each persona’s captions $\{c_i\}$ are embedded using \texttt{text-embedding-3-small}~\cite{openai2024gpt4technicalreport},
and cosine similarities $s_i = \mathbf{q}^\top \mathbf{v}_i$ are computed with the query $p$.
Weights are normalized as $w_i = \frac{\exp(s_i/\tau)}{\sum_j \exp(s_j/\tau)},$ where $\tau$ is a softmax temperature controlling retrieval sharpness.
This Boltzmann weighting represents the \emph{maximum-entropy solution}
for expected semantic alignment under a temperature constraint~\cite{PhysRev.106.620},
guaranteeing smooth attention while avoiding brittle hard cutoffs. 

\textit{Entropy Guided post Selection.}
We then measure entropy $H = -\sum_i w_i \log w_i, \quad n_{\text{eff}} = \exp(H),$ where $n_{\text{eff}}$ approximates the \emph{effective number of relevant posts},a theoretically grounded proxy for query specificity.
Broader prompts (e.g., ``in the park'') yield higher $H$ and therefore encourage more diverse retrieval,
whereas narrower ones (e.g., ``in Kashmiri traditional dress'') produce lower entropy, focusing the selection.
To balance adaptivity and efficiency, we cap the retrieved posts given the budget $Q$ (total number of visual elements to sample from categories $\mathcal{C}=\{\text{fg}, \text{bg}, \text{lighting}, \text{pose}, \ldots\}$), set as $K = \min\!\left( \left\lfloor n_{\text{eff}} \right\rfloor,\; 2 \times  Q \right)$, ensuring controlled expansion without over-retrieval for broad prompts.

\textit{Quota Allocation.} Each post contributes elements from categories $\mathcal{C}$. Given category $c' \in \mathcal{C}$, we allocate quotas to each post $i$ as: $q_i^{(c')} = \left\lfloor 
\frac{w_i \cdot n_i^{(c')}}{\sum_j w_j \cdot n_j^{(c')}} \cdot Q_{c'}
\right\rfloor$
where $n_i^{(c')}$ is the number of available elements in category $c'$ for post $i$, and $Q_{c'}$ is the total budget for category $c'$. Remainders are allocated to largest–fraction posts. This rule ensures the proportional-fair allocation objective so that high-weight posts get more samples, but low-weight ones still contribute diversity.

\textit{\textbf{Element-level retrieval.}}
Within the top-$K$ posts we prioritize the categories based on the prompt $p$ using semantic relevance $\text{score}_{k} = \cos(\phi(\mathbf{c}_k),\, \phi(p)),$ ($\phi$ is a lightweight transformer encoder (\texttt{MiniLM})~\cite{MiniLM}). Within each category, elements are ranked based on the closeness to the $p$ using the same \texttt{MiniLM}~\cite{MiniLM}, and the top-$q_i^{(k)}$ are selected.

\paragraph{Prompt Composition.} The selected elements $\mathcal{E}_p$ are concatenated with persona summary $\mathcal{S}_p$
into $p' = f_{\text{compose}}(p,\mathcal{S}_p,\mathcal{E}_p,L)$ under a token-length budget $L$.
This yields a re-prompt enriched with stylistic and contextual cues consistent with the persona’s memory. Based on the budget, $f_{\text{compose}}$, can be an LLM refining the story arc for the generation or a simple text concatenation.

\subsection{Learnable Feedback Simulation}
\label{sec:feedback}

While VPRAG uses persona aligned retrieval, personalization also involves subjective preference learning. We therefore introduce a small learnable feedback module
to approximate user-specific value functions.
Given persona $\mathcal{P}$ with subjective preferences and generated prompt $p'$,
a vision–language judge (VLM) outputs an alignment score $s_{\text{VLM}}\!\in\![0,1]$.
We train a cross-attention predictor $f_\theta$
to estimate $\hat{s}_{\text{VLM}} = f_\theta(\text{Emb}(p'),\text{Emb}(\mathcal{P}))$,
and re-rank candidates by $p'^* = \arg\max_m f_\theta(\text{Emb}(p'_m),\text{Emb}(\mathcal{P}))$.
We use this component as a smaller scale proof of concept to encourage future extensions of VPTT Framework toward closed-loop personalization.

\subsection{VPTT Score: A Differentiable Proxy for Personalization}
\label{sec:pgp}

We now introduce $\mathrm{VPTT_{score}}$, a quantitative metric that serves as the text-level scalable foundation for the VPTT triangle and a convex surrogate of the personalization objective in Eq.~\ref{eq:vptt}.  
$\mathrm{VPTT_{score}}$ combines four interpretable metrics that jointly approximate alignment, fidelity, and originality:
\textit{Persona Alignment (PA)}, \textit{GS Reconstruction (GS)}, 
\textit{Cluster Proximity (CP)}, and \textit{Novelty (NV)}.

\textbf{(1) Persona Alignment (PA).}
This term measures semantic coherence between the generated prompt $p'$  
and the textual description of the persona $\mathcal{P}$: $\text{PA}(p', \mathcal{P}) =
\cos\!\big(\text{Emb}(p'),\, \text{Emb}(\mathcal{P})\big)$.  

\textbf{(2) GS Reconstruction (GS).}
To measure content fidelity, we represent each persona’s caption embeddings $\{v_i\}$  
as an orthonormal basis $B$ using the Gram–Schmidt process.  
For a generated prompt embedding $v_p$,  $\text{GS}(p', C) = \cos\!\big(v_p,\, B(B^\top v_p)\big)$ which evaluates how well $p'$ can be reconstructed from the assets’s semantic span. GS measures subspace fidelity i.e. whether a generation stays within the semantic manifold defined by the persona’s gallery rather than mere pairwise similarity.

\textbf{(3) Cluster Proximity (CP).} To assess thematic consistency, all asset captions are clustered in the GS basis thematic centroids $\{c_k\}$.  
The hard version used for evaluation is $\text{CP}(p', C)
= \exp\!\big(-\min_k \|v_p' - c_k\|_2 \big),$ while the differentiable relaxation replaces $\min$ with a temperature-controlled softmin: $\widetilde{\text{CP}}(p', C)
= \sum_k 
\frac{\exp(-\|v_p'-c_k\|_2/\tau)}{\sum_j \exp(-\|v_p'-c_j\|_2/\tau)}.$

\textbf{(4) Novelty (NV).}
Novelty penalizes verbatim reuse of retrieved captions.  
The discrete version measures maximum trigram overlap: $\text{NV}(p',C)
= 1 - \max_i 
\frac{|\text{Tri}(p') \cap \text{Tri}(c_i)|}{|\text{Tri}(p')|}.$ For differentiable analysis, we define a soft-overlap relaxation:
$
\widetilde{\text{NV}}(p', C)
= 1 - 
\max_i 
\frac{
\sum_t 
\cos(\phi_t(p'),\, \phi_t(c_i))
}{
|\text{Tri}(p')|
},
$
where $\phi_t(\cdot)$ denotes continuous n-gram embeddings (via small sentence transformer for example \texttt{MiniLM}~\cite{MiniLM}).  

\vspace{-0.3cm}
\paragraph{Combined Score.}

The overall proxy is a convex weighted combination: $\mathrm{VPTT_{score}}
= 0.20\,\text{PA}
+ 0.30\,\text{GS}
+ 0.30\,\text{CP}
+ 0.20\,\text{NV}.$
Empirically, GS and CP correlate most strongly with human visual fidelity,  
so we assign them higher weight ($0.3$ each).  
PA measures semantic alignment ($0.2$),  
while NV promotes originality and prevents overfitting ($0.2$).  
The weighting satisfies $\sum_i \lambda_i = 1$,  
forming an unbiased convex estimator of $\mathcal{J}$. For tasks with limited prompt budgets (e.g., adding exactly three retrieved phrases), the novelty term becomes less meaningful as textual overlap is bounded by design. We therefore use the normalized variant $\mathrm{VPTT_{score}\text{-}c} 
= \tfrac{1}{3}(\text{PA} + \text{GS} + \text{CP}),$ which equally weighs the three active components. We further justify the weights in Sec~\ref{sec:metrics} while computing the correlations. The novelty term is also set to zero for the baselines not conditioned on the captions. While our experiments report the hard (evaluation) forms for interpretability,
the differentiable variant makes $\mathrm{VPTT_{score}}$ suitable as a learnable objective in future
personalization pipelines.

\vspace{-0.2cm}

\section{Evaluations}

\subsection{Baselines}
\label{sec:baselines}

We benchmark VPTT Framework against two baseline categories.  
First, scalable privacy-safe pipelines including \emph{Baseline} - no access to any asset, \emph{Persona Only} - access to demographics information, and Baseline RAG \emph{BRAG}~\cite{chen2024tailoredvisionsenhancingtexttoimage}, a strong baseline with access to all the persona captions for personalization (see Figure.~\ref{fig:pipeline}). These operate via retrieval and rewriting without model retraining, allowing large-scale evaluation across $10{,}000$ personas.  
Second, we reference high-cost personalization baselines such as DB-LoRA~\cite{dblora}, Flux~\cite{FLUX}, DrUM~\cite{kim2025drawmindpersonalizedgeneration}, MLLM~\cite{openai2024gpt4technicalreport, yang2025qwen3technicalreport}, and VIPER~\cite{ViPer}, which rely on user-specific fine-tuning or only preference optimization.  
These are computationally intensive and non-scalable, so we evaluate them only on smaller subsets and report results in the Supplementary. This separation highlights VPTT’s focus on scalable, privacy-safe personalization while remaining comparable to existing high-fidelity methods.

\subsection{Quantitative Evaluation}

Evaluating the VPTT is intrinsically challenging because the outcome depends on a cascade of interacting systems:
\begin{enumerate}[label=\arabic*), nolistsep, topsep=0pt, leftmargin=*]
    \item \textbf{Prompt Generation:} The rewriter LLM must faithfully express a persona’s stylistic intent.  
    \item \textbf{Image Generation:} The T2I or I2I model must accurately translate those prompts into coherent visual content.  
    \item \textbf{Evaluation:} The VLM judge must perceive the subtle consistency between the generated content and the persona’s authentic visual identity.  
\end{enumerate}

VPTT performance improves as these three domains mature. To systematically evaluate them, we design a three-stage protocol addressing three central questions (\textbf{Q1–Q3}). All experiments are conducted across a spectrum of models from open-source \textbf{Qwen2.5-7B-Instruct}~\cite{yang2025qwen3technicalreport} to efficient \textbf{GPT-4o-mini}~\cite{openai2024gpt4ocard} and high-capacity \textbf{Gemini-2.5-Pro}~\cite{comanici2025gemini25pushingfrontier} ensuring robustness across compute budgets. To make the evaluation holistic, we consider both image generation and editing tasks.

\subsubsection{Q1: Can We Trust Our Metrics?}
\label{sec:metrics}
Before scaling the evaluation, we verify that our automated metrics i.e. VLM judgment and the text-only $\mathrm{VPTT_{score}}$ faithfully approximate human perception.

\paragraph{Human Study.}
We collected about 6,000 human ratings using images across \textit{four} methods (see Table.~\ref{tab:generation_scores}), \textit{three} LLM generations and \textit{two} tasks (image generation ``A preferred outdoor spot" and editing ``Here is a convention center. Add a preferred event"), from 20 annotators. Inter-annotator agreement was substantial (Kendall’s $W = 0.651  \pm0.141$ for Generation, $0.564 \pm 0.209$ for Editing), confirming consistent human understanding of personal authenticity.

\vspace{-0.5cm}
\paragraph{Metric Calibration and Validity.}

We validate the proposed metrics by measuring Spearman’s rank correlation ($\rho$) between automated judgments and human ratings (Figure~\ref{fig:vptt}).  
For efficient evaluation, we use 10 visually and semantically matched posts out of 30 under a budgeted evaluation setup (Sec.~\ref{sec:pgp}). We calibrate the VLM judges using GPT-4o and Gemini-2.5-Pro, wherever applicable to remove evaluation bias on a small set. In evaluation of the whole set, VLM-based judgments strongly align with human perception (combined $\rho = 0.67$, generation: $0.75$).  
Our text-only $\mathrm{VPTT_{score}\text{-}c}$ metric achieves comparable agreement (combined $\rho = 0.68$, generation: $0.78$) with a Top-2 agreement accuracy of 99\%, confirming its reliability as a human-perceptual proxy.  
$\mathrm{VPTT_{score}\text{-}c}$ also correlates well with VLM scores (combined $\rho = 0.57$, generation: $0.70$), indicating consistent cross-modal alignment.  
While editing correlations are lower ($\rho \approx 0.5$) due to the finer granularity of localized visual edits and potential perceptual losses after downsampling, generation consistently exceeds $0.7$, demonstrating the robustness of our metric design. Finally, we report the averaged raw scores in Table.~\ref{tab:generation_scores} where our method VPRAG is a clear winner across all the evaluations. Overall, these results establish $\mathrm{VPTT_{score}\text{-}c}$ as a fast, low-cost, and perceptually grounded surrogate for human evaluation in large-scale personalization studies.

\begin{table}[t]
\centering
\captionsetup{font=small}
\caption{Quantitative comparison for generation and Editing Tasks across 6000 human annotations.
We report mean (\textbf{Avg.}) and accuracy (\textbf{Acc.}) scores for three evaluation levels:
text‐based \textbf{$\mathrm{VPTT_{score}\text{-}c}$ (0–1)}, vision‐language VLM (0–5), and human judgments Human (0–5). Higher is better for all.}
\label{tab:generation_scores}
\scriptsize
\setlength{\tabcolsep}{4pt}
\renewcommand{\arraystretch}{1.05}
\begin{tabular}{lcccccc}
\toprule
\multirow{2}{*}{\textbf{Method}} &
\multicolumn{2}{c}{$\mathrm{VPTT_{score}\text{-}c}$ (Text)} &
\multicolumn{2}{c}{VLM (Visual)} &
\multicolumn{2}{c}{Human (Perceptual)} \\
\cmidrule(lr){2-3} \cmidrule(lr){4-5} \cmidrule(lr){6-7}
& \textbf{Avg.} & \textbf{Acc.} & \textbf{Avg.} & \textbf{Acc.} & \textbf{Avg.} & \textbf{Acc.} \\
\midrule
Baseline               &  0.329     & 	0.0\%  &    	2.41    &  	4.6\%     & 1.64 &  0.70\%\\
Persona Only              &  0.400      &  	7.3\%     &  	3.32      & 19.2\%    & 2.51 &  16.0\% \\
BRAG                    &  0.420      &  	19.3\%  & 3.52       &  	21.6\%   & 2.69 & 21.3\%  \\
VPRAG (Ours)          & \textbf{0.464}       &  \textbf{73.3\%}    &  \textbf{4.32}      & \textbf{54.6\%}    & \textbf{3.34} & \textbf{62.0\%}   \\
\bottomrule
\end{tabular}
\vspace{-0.5cm}
\end{table}


\begin{table}[t]
\centering
\captionsetup{font=small}
\caption{Comparison of Generation and Editing tasks on 200 personas after VLM calibration across 3 LLM rewrite methods. We report mean $\mathrm{VPTT_{score}\text{-}c}$ (V-c) and VLM scores along with wining accuracy (\%). Higher is better.}
\label{tab:gen_edit}
\vspace{-0.5em}
\scriptsize
\setlength{\tabcolsep}{3pt}
\renewcommand{\arraystretch}{1.05}

\begin{tabular}{lcccccccc}
\toprule
\multirow{2}{*}{\textbf{Method}} &
\multicolumn{4}{c}{\textbf{Generation}} &
\multicolumn{4}{c}{\textbf{Editing}} \\
\cmidrule(lr){2-5} \cmidrule(lr){6-9}
& V-c & Acc. & VLM & Acc. &
 V-c & Acc. & VLM & Acc. \\
\midrule
Baseline     & 0.343 & 0.0\%  & 2.21 & 1.4\%  & 0.322 & 0.0\%  & 2.97 & 10.5\% \\
Persona Only      & 0.402 & 1.2\%  & 2.98 & 5.9\%  & 0.399 & 9.2\%  & 3.44 & 18.5\% \\
BRAG         & 0.451 & 18.4\% & 4.04 & 25.6\% & 0.415 & 15.3\% & 3.75 & 24.3\% \\
VPRAG (Ours) & \textbf{0.472} & \textbf{41.7\%} & \underline{4.08} & \underline{31.0\%} &
\textbf{0.448} & \textbf{47.2\%} & \textbf{4.03} & \textbf{30.8\%} \\
Comb. (Ours) & \underline{0.472} & \underline{38.8\%} & \textbf{4.30} & \textbf{36.1\%} &
\underline{0.436} & \underline{28.3\%} & \underline{4.03} & \underline{15.8\%} \\
\bottomrule
\end{tabular}
\vspace{-1.5em}
\end{table}

\begin{table*}[t]
\centering
\caption{Main text-level results across 10,000 personas and three LLM models. We report the novelty-adjusted $\mathrm{VPTT_{score}}$ (\textbf{V}), plus Cohen's $\textbf{d}$~\cite{cohen2013statistical} ($\textbf{d} = \frac{|\mu_{\text{best}} - \mu_{\text{method}}|}{s_{\text{pooled}}}$), measuring effect size relative to the best-performing method per row ($\mu_{\text{best}}$) across 20,000 samples per entry. \textbf{Bold} indicates the best method and \underline{underline} the second-best. The \emph{Baseline} and \emph{Persona Only} methods consistently underperform across both generation and editing tasks. Our \emph{VPRAG} and \emph{Comb.} (BRAG + VPRAG) methods achieve the best overall performance, with \emph{Comb.} performing slightly better for 4o-mini (GPT-4o-mini~\cite{openai2024gpt4technicalreport}) and Gemini (Gemini-2.5-pro~\cite{comanici2025gemini25pushingfrontier}), while \emph{VPRAG} excels for Qwen (Qwen2.5-7B-Instruct~\cite{yang2025qwen3technicalreport}). Higher Cohen's $\textbf{d}$ values ($d \geq 0.5$ indicates medium to large effects) demonstrate substantial performance differences, particularly between persona-based methods and baselines. See supplementary material for detailed score breakdowns.}
\label{tab:pgp_models_methods}
\scriptsize
\setlength{\tabcolsep}{3.5pt}
\renewcommand{\arraystretch}{1.05}
\vspace{-0.3cm}
\begin{subtable}[t]{0.49\textwidth}
\centering
\caption{Generation}
\label{tab:pgp_models_methods_gen}
\begin{tabular}{l*{10}{r}}
\toprule
& \multicolumn{2}{c}{Baseline} & \multicolumn{2}{c}{Persona Only} & \multicolumn{2}{c}{BRAG} & \multicolumn{2}{c}{VPRAG} & \multicolumn{2}{c}{Comb.} \\
\cmidrule(lr){2-3}\cmidrule(lr){4-5}\cmidrule(lr){6-7}\cmidrule(lr){8-9}\cmidrule(lr){10-11}
\textbf{Model} & \textbf{V} & \(\boldsymbol{d}\) & \textbf{V} & \(\boldsymbol{d}\) & \textbf{V} & \(\boldsymbol{d}\) & \textbf{V} & \(\boldsymbol{d}\) & \textbf{V} & \(\boldsymbol{d}\) \\
\midrule
Qwen &0.316  &11.9 &0.389  &8.3 &0.581  &1.1 &\textbf{0.631}  &NA &\underline{0.602}  &0.7  \\
4o-mini &0.316  &12.6 &0.402  &8.4 &0.628  &0.5 &\underline{0.640}  &0.1 &\textbf{0.644}  &NA  \\
Gemini &0.316  &9.8 &0.379  &7.1 &0.616  &0.3 &\underline{0.625}  &0.2 &\textbf{0.632}  &NA  \\
\bottomrule
\end{tabular}
\end{subtable}\hfill
\begin{subtable}[t]{0.49\textwidth}
\centering
\caption{Editing}
\label{tab:pgp_models_methods_edit}
\begin{tabular}{l*{10}{r}}
\toprule
& \multicolumn{2}{c}{Baseline} & \multicolumn{2}{c}{Persona Only} & \multicolumn{2}{c}{BRAG} & \multicolumn{2}{c}{VPRAG} & \multicolumn{2}{c}{Comb.} \\
\cmidrule(lr){2-3}\cmidrule(lr){4-5}\cmidrule(lr){6-7}\cmidrule(lr){8-9}\cmidrule(lr){10-11}
\textbf{Model} & \textbf{V} & \(\boldsymbol{d}\) & \textbf{V} & \(\boldsymbol{d}\) & \textbf{V} & \(\boldsymbol{d}\) & \textbf{V} & \(\boldsymbol{d}\) & \textbf{V} & \(\boldsymbol{d}\) \\
\midrule
Qwen &0.306  &12.0 &0.378  &8.7 &0.583  &1.1 &\textbf{0.626}  &NA &\underline{0.586}  &1.0  \\
4o-mini &0.306  &12.0 &0.384  &8.8 &0.596  &0.9 &\textbf{0.626}  &NA &\underline{0.610}  &0.5  \\
Gemini &0.306  &10.7 &0.372  &8.1 &0.583  &0.6 &\underline{0.605}  &0.0 &\textbf{0.606}  &NA  \\
\bottomrule
\end{tabular}
\end{subtable}
\vspace{0.25em}
\end{table*}

\begin{figure*}[t] 
 \vspace{-0.2cm}
    \centering \includegraphics[width=1.0\linewidth]{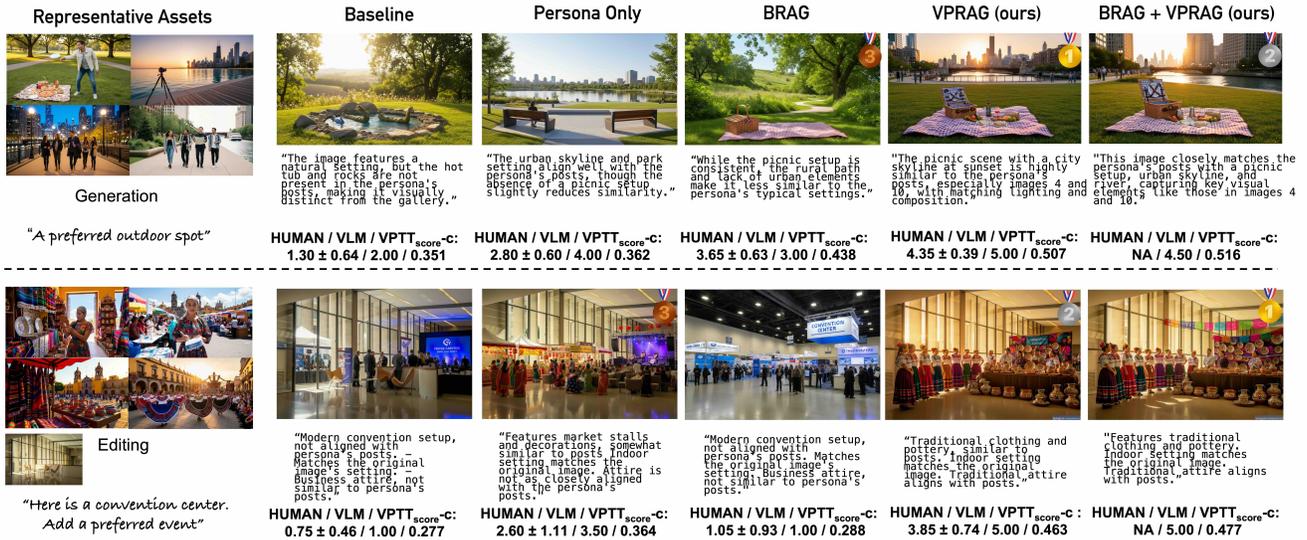}
    \vspace{-0.7cm}
    \caption{\textbf{Qualitative Comparison across Generation and Editing Tasks.}
Representative examples from the VPTT-Bench showing outputs from five methods: Baseline, Persona Only, BRAG, VPRAG (ours), and BRAG + VPRAG (ours). Each sample is evaluated using human, VLM (reasoning shown), and text-level $\mathrm{VPTT_{score}\text{-}c}$  scores, where higher indicates closer alignment to the persona’s assets. Our methods achieve the highest perceptual and text–visual consistency, confirming effective contextual personalization.}
    \label{fig:comp1}
    \vspace{-0.3cm}
\end{figure*}

\subsubsection{Q2: Does a Better Prompt Create a Better Image?}

With calibrated evaluators, we conduct the main VPTT experiment on 200 personas on \textit{two} tasks ( across \textit{three} LLM models and \textit{five} methods) under a fixed “three-phrase budget” to ensure fair comparison. This part disentangles what visual generation is able to achieve with models' ability to generate authentic detailed prompts (we evaluate that next). Evaluation of this extended dataset mirrors the correlation $\rho = 0.53$ (generation : $0.66$) of the previous section. Table.~\ref{tab:gen_edit} shows the results (averaged across LLMs) for the generation and editing tasks. The evaluation again shows how hierarchical controllable retrieval does not confuse the models and produce better alignments.

\subsubsection{Q3: Is the Architecture Robust at Scale?}

To assess generalization, we evaluate all models text-only across our entire VPTT-Bench benchmark of 10,000 personas and four tasks (\textit{two} generation, \textit{two} editing , see Figure.~\ref{fig:main}), totaling 120,000 prompt evaluations. The prompts are limited to 150 words and a budget of 3 is allocated to all visual element Categories $\mathcal{C}$. The elements are arranged in decreasing order of relevance and LLM is given freedom to choose from the list to orchestrate a story arc. As shown in Table~\ref{tab:pgp_models_methods}, naive rewriters (BRAG) overfit to captions (often copy-pasting them), earning high alignment but low originality scores (more detailed in Supplementary) and hence falling short. In contrast, VPRAG consistently achieves the best composite $\mathrm{VPTT_{score}}$, maintaining the optimal balance between alignment and originality across all rewriter backbones. This large-scale experiment demonstrates that VPRAG scales linearly, generalizes across models, and sustains perceptual authenticity without retraining.

\subsubsection{Downstream Study: Feedback Simulation }

We evaluate feedback simulation on a smaller subset of 200 personas (10,000 labeled examples) as a proof of concept rather than a core benchmark. Although this component is not used in our main quantitative evaluations, it demonstrates that compact models can learn to simulate user-level preference alignment from limited supervision. We sample diverse simulated profiles (95\% occupation uniqueness, 96 countries, 10 ethnicity groups) and use GPT-4o~\cite{openai2024gpt4ocard} to generate 50 labeled prompts per profile, 20 aligned, 20 misaligned, and 10 neutral, yielding 10,000 labeled examples with profile-level splits (130/20/50 train/val/test). A compact cross-attention regressor (128-dim, 4 heads) achieves 73.8\% overall accuracy (MAE: 0.1259) and 91.6\% accuracy on aligned preference predictions for 50 unseen users (2{,}525 prompts), with only a 0.7\% validation–test gap, showing that compact models can effectively capture persona-aware preferences while generalizing to new users. We leave large scale studies to future extensions.

\subsection{Qualitative Results}

VPRAG produces visually coherent and persona-faithful generations across diverse profiles. 
By retrieving fine-grained visual cues such as lighting, attire, scene semantics, and stylistic markers, VPRAG enriches the composed prompts while preserving originality and user-specific visual elements (Figure~\ref{fig:main}). 
These examples also highlight VPRAG’s ability to perform cross model personalization, where VPRAG produces consistent personalization across QWEN-Image~\cite{yang2025qwen3technicalreport}  and Nano-Banana~\cite{nano_banana}.

Compared to the persona-only baseline (Figure~\ref{fig:comp1}) and the BRAG baseline, VPRAG achieves stronger contextual grounding, sharper visual fidelity, and more consistent preservation of persona style. 
For editing tasks, it additionally injects semantically relevant visual elements. Results for remaining baselines and additional profiles are provided in the Supplementary.
\vspace{-0.1cm}
\section{Conclusion}
We introduced the Visual Personalization Turing Test (VPTT) as a principled paradigm for evaluating contextual visual personalization, and proposed the VPTT Framework, a scalable system that operationalizes this paradigm. The framework integrates VPTT-Bench, the VPRAG retrieval engine, and the $\mathrm{VPTT_{score}}$ metric into a closed-loop pipeline for simulation, generation, and evaluation without any per-user retraining. Our results show strong alignment among human judgments, VLM judges, and the text-only $\mathrm{VPTT_{score}}$, validating the framework as an efficient, privacy-safe foundation for personalized generative models. Future work will incorporate opt-in and federated real-user signals to further bridge simulated and real personalization while preserving user privacy.
\clearpage
\setcounter{page}{1}
\maketitlesupplementary
\begin{figure*}[t] 
    \centering \includegraphics[width=1.0\textwidth]{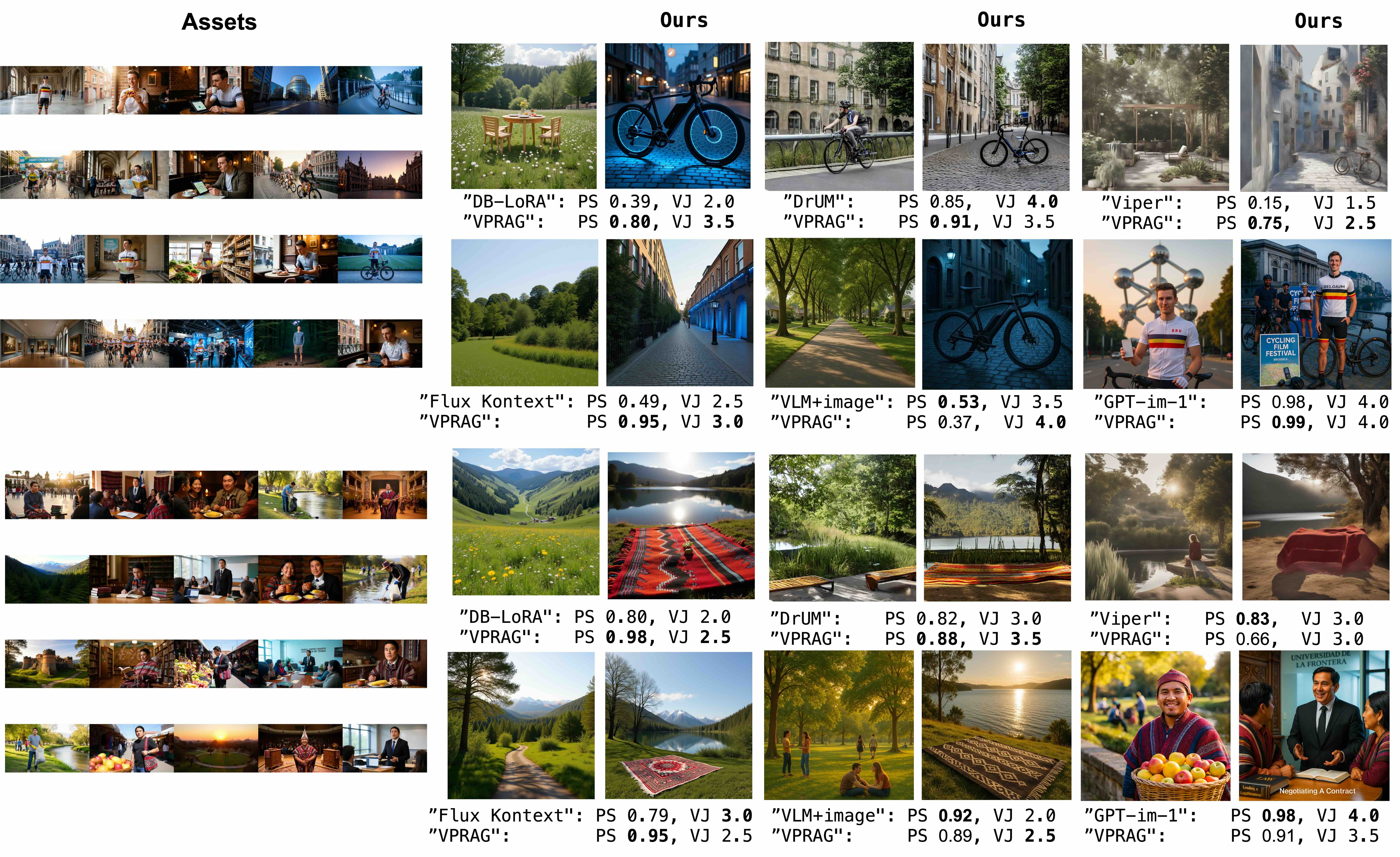}
    \caption{\textbf{Comparison to Visual Baselines.}
We compare VPRAG (along three columns) against a broad set of visual personalization baselines, including fine-tuning approaches, preference-driven personalization methods, and multimodal LLM (MLLM)–based in-context techniques.  
Evaluation is conducted using two metrics: the VIPER Proxy Score (PS)~\cite{ViPer} and the Gemini VLM Judge (see Sec. 4 in the main paper).  
Across more challenging and nuanced examples shown in the figure, VPRAG consistently emerges as an efficient and controllable personalization method, performing on par with or outperforming these substantially more expensive baselines.}
    \label{fig:visual}
\end{figure*}

\begin{figure}[t] 
    \centering \includegraphics[width=1.0\linewidth]{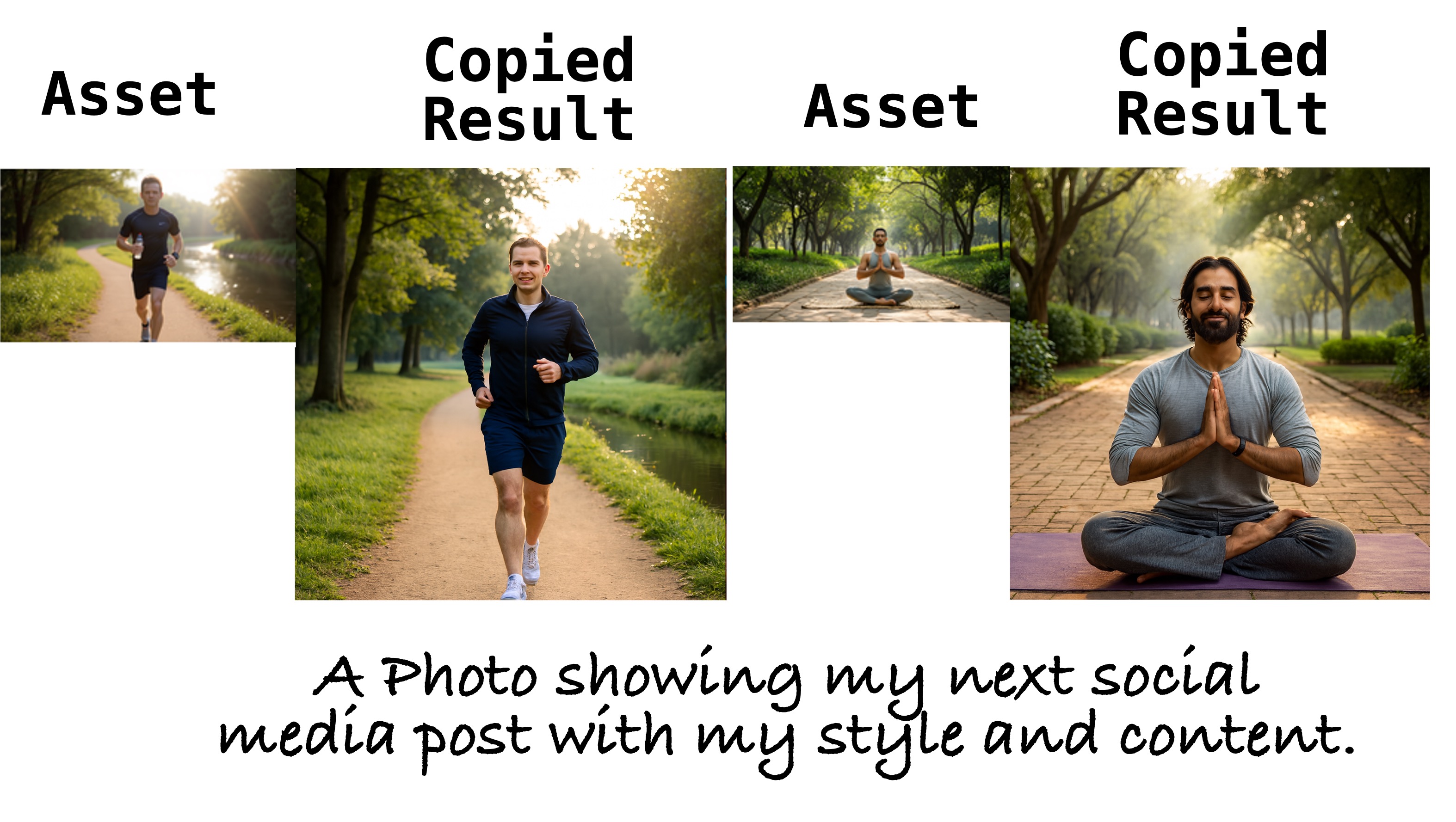}
    \caption{\textbf{Copy-Paste Effect.} The baselines including MLLMs suffer from copy-paste effect where the generations and edits only consider a single or few images of the user assets. }
    \label{fig:copying}
\end{figure}

\begin{figure*}[t] 
    \centering \includegraphics[width=1.0\textwidth]{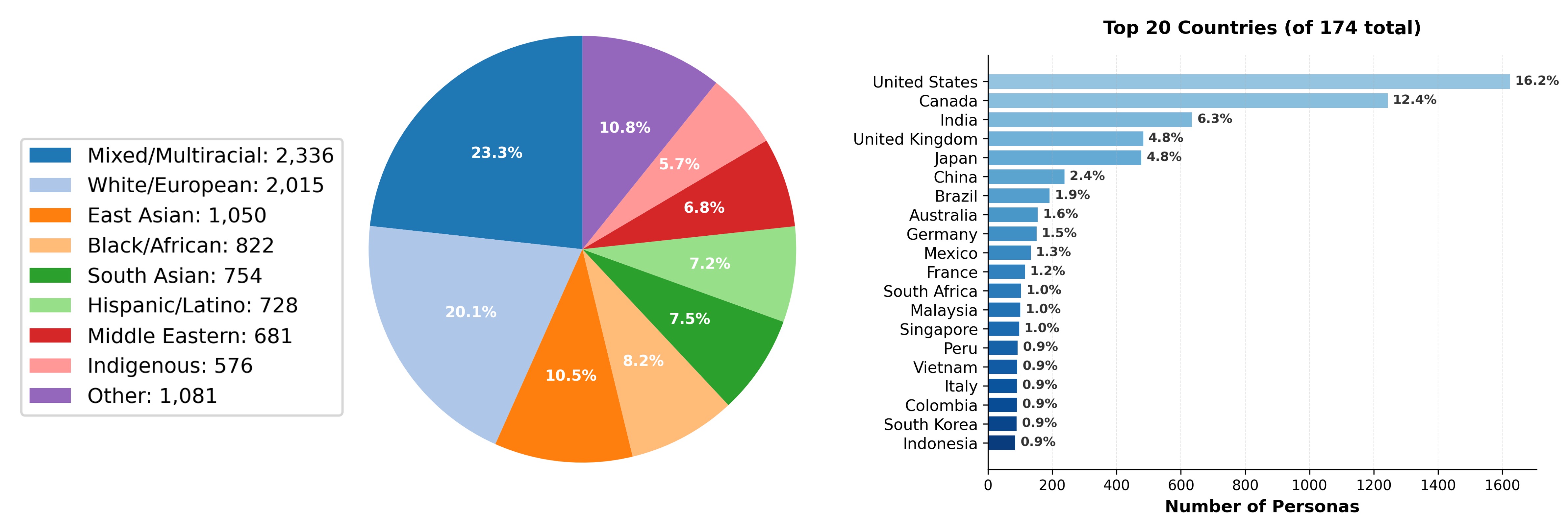}
    \caption{\textbf{VPTT-Bench Ethnicity and Location Diversity}. Ethnicity and location diversity of the users in VPTT-Bench}
    \label{fig:diversity1}
\end{figure*}

\begin{figure}[t] 
    \centering \includegraphics[width=1.0\linewidth]{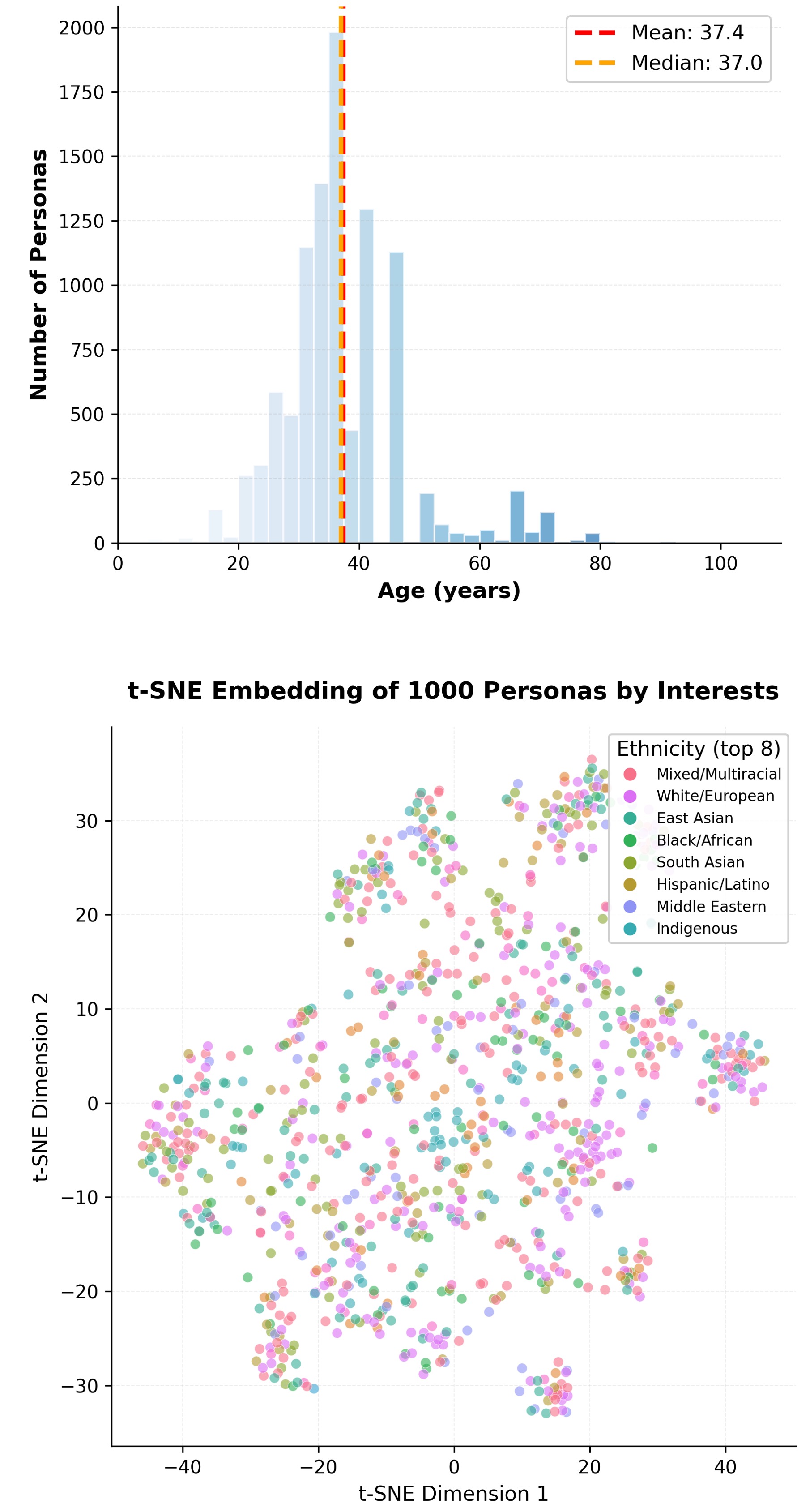}
    \caption{\textbf{VPTT-Bench Age and Interest Diversity}. Age Distribution and t-sne visualization (interests) of first 1000 users.}
    \label{fig:diversity2}
\end{figure}

\begin{figure*}[t] 
    \centering \includegraphics[width=1.0\linewidth]{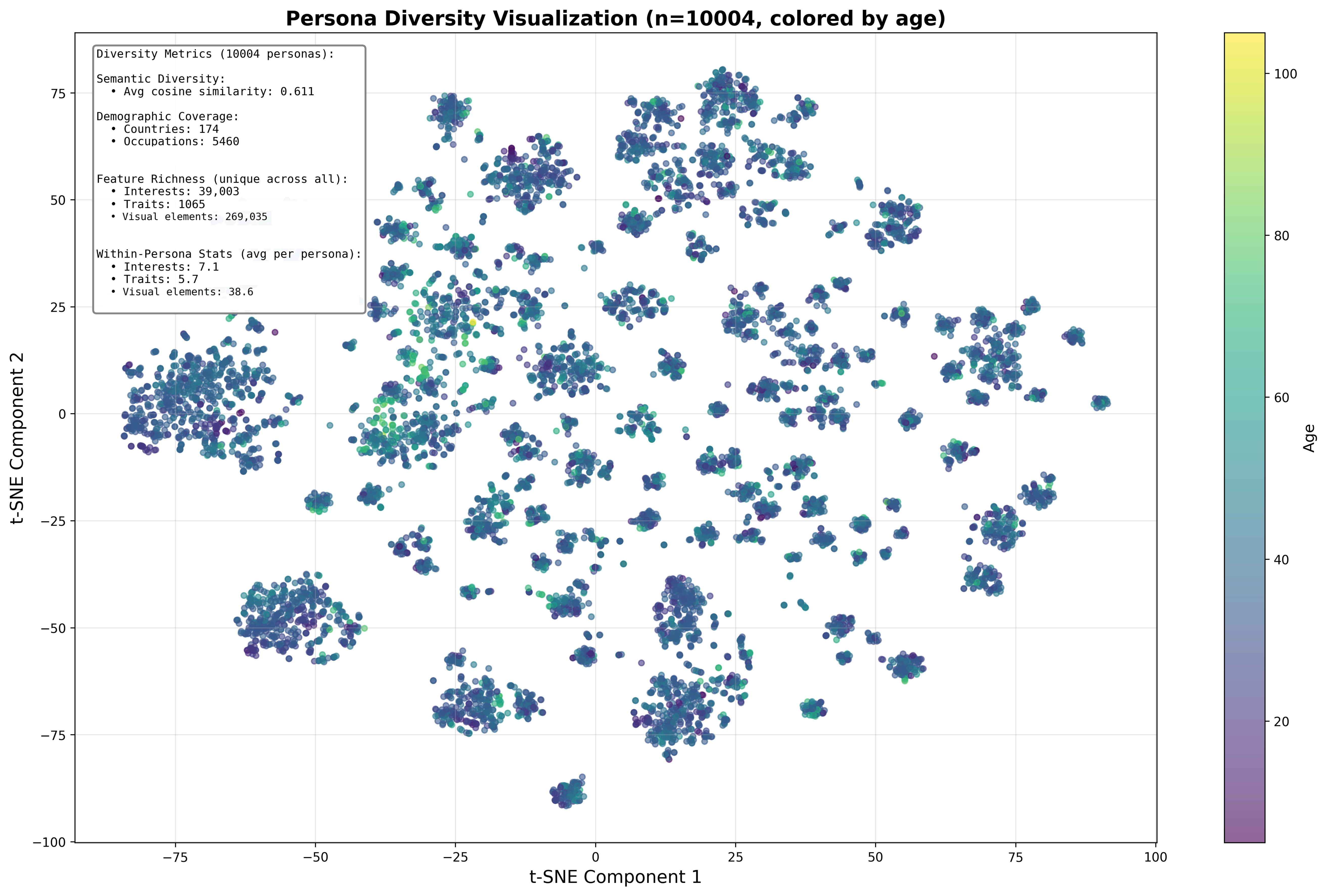}
    \caption{\textbf{Diversity of 10K Synthetic Personas.} We visualize the diversity of our 10,004 synthetic personas using t-SNE dimensionality reduction on averaged caption embeddings (OpenAI text-embedding-3-small, 1536-dim) from each persona's 30-image gallery. Points are colored by age. The average pairwise cosine similarity of 0.611 indicates balanced diversity ; personas occupy a shared human aesthetic space while maintaining distinct individual preferences. Our dataset spans 174 countries and 5,460 unique occupations, with 39,003 unique interests and 269,035 visual elements across all personas. Each persona averages 7.1 interests, 5.7 personality traits, and 38.6 visual elements, ensuring rich and diverse personalization signals for image generation models.}
    \label{fig:diversity3}
\end{figure*}

\begin{figure*}[t] 
    \centering \includegraphics[width=1.0\textwidth]{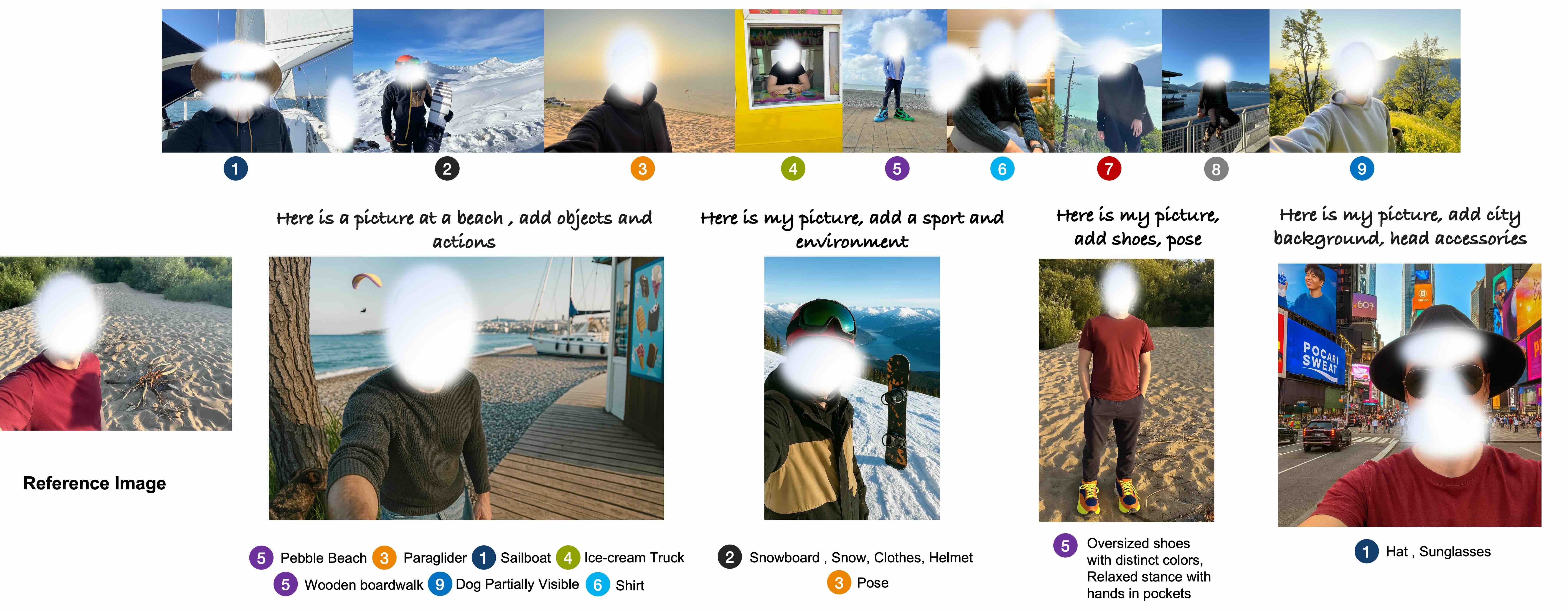}
    \caption{This Figure is only for illustration and not a part of the main dataset. The human figures shown in the sample images are non-author volunteers who provided consent. Their faces and all identifying cues (e.g., location) are fully anonymized.}
    \label{fig:user1}
\end{figure*}

\begin{figure*}[t] 
    \centering \includegraphics[width=1.0\textwidth]{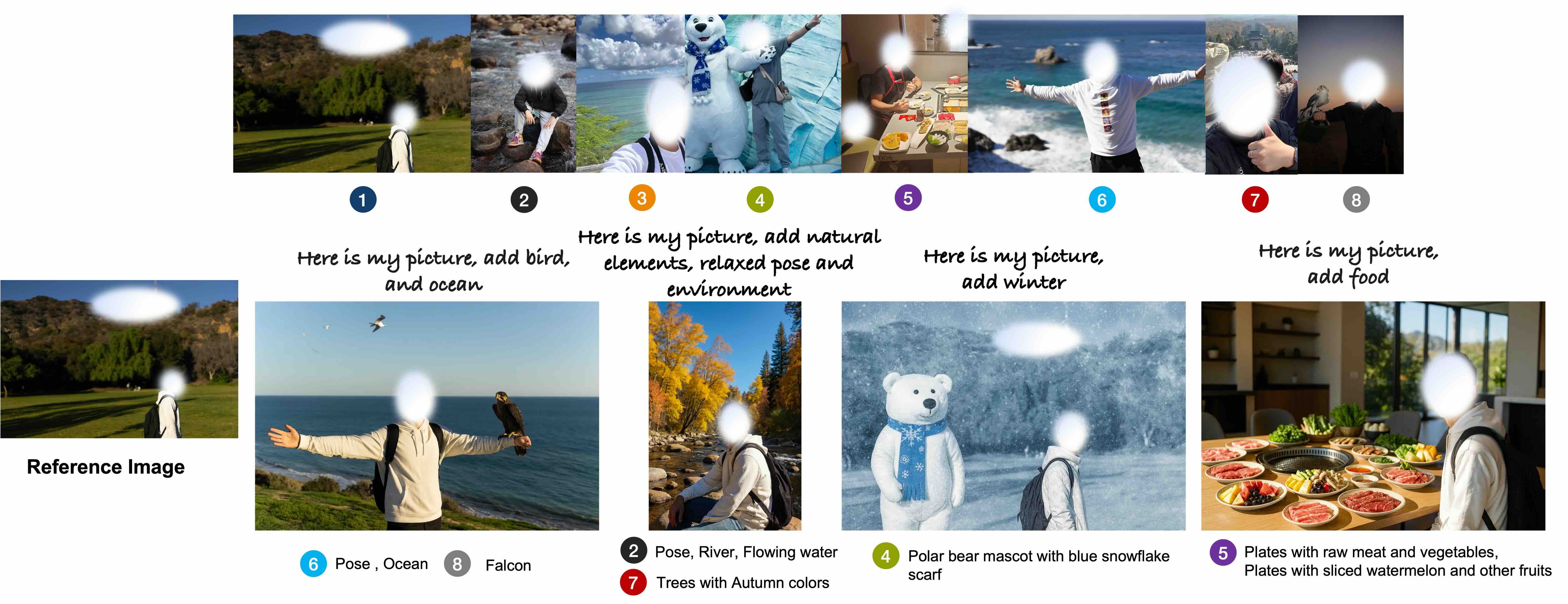}
    \caption{ This Figure is only for illustration and not a part of the main dataset. The human figures shown in the sample images are non-author volunteers who provided consent. Their faces and all identifying cues (e.g., location) are fully anonymized.}
    \label{fig:user2}
\end{figure*}

\section*{S.1 Additional Details: Formalization of the VPTT Evaluation Protocol}

This section provides additional mathematical clarification of the Visual
Personalization Turing Test (VPTT) evaluation protocol described in the main
paper. The formalization offers a
rigorous scientific grounding for the task and mitigates subjective interpretation.

\paragraph{Setup.}
A persona is defined as $P = \{d, E, C\}$ (demographics, structured visual
elements, and caption memory). Given a query $p$, the personalization system
produces a rewritten prompt $p'$ and a generated visual output
\[
X = \mathcal{G}(p') \in \mathcal{X},
\]
where $\mathcal{G}$ denotes the visual generative model (see
Sec.3.2, main paper). For brevity, we write
$X \sim G(\cdot \mid p,P)$ to denote the overall personalization pipeline that
includes retrieval, prompt rewriting, and generation.

\paragraph{Judge function.}
As described in the main paper, VPTT evaluates whether $X$ is
\emph{indistinguishable from content that the persona might plausibly create or
share}. We formalize this via a judge function
\[
J : \mathcal{X} \times \mathcal{P} \rightarrow [0,1],
\quad J(X,P) = \text{plausibility score}.
\]
Human annotators and VLM-based judges provide plausibility judgements on a 0--5 Likert scale, which
can be linearly normalized to $[0,1]$.
In large-scale evaluations, we substitute $J$ with the
$\mathrm{VPTT_{score}}$ (Sec.~4, main paper), which serves as a scalable proxy.

\paragraph{Expected VPTT performance.}
Let $\mu$ be the distribution over persona–query pairs. The expected VPTT
performance of a generator $G$ is
\begin{equation}
\Pi(G)
=
\mathbb{E}_{(P,p) \sim \mu}
\Big[
\mathbb{E}_{X \sim G(\cdot \mid p,P)}
\big[
J(X,P)
\big]
\Big].
\tag{S1}
\end{equation}

\paragraph{Finite-sample estimator.}
Using $N$ personas and $K$ queries per persona, we estimate Eq.~(S1) as
\begin{equation}
\widehat{\Pi}(G)
=
\frac{1}{N K}
\sum_{i=1}^{N}
\sum_{j=1}^{K}
J\!\left( X_{ij}, P_i \right),
\quad
X_{ij} \sim G(\cdot \mid p_{ij}, P_i).
\tag{S2}
\end{equation}

\paragraph{Judging modalities.}
\begin{itemize}
\item \textbf{Human judges:} $J(X,P)$ is the mean normalized Likert score over
annotators.
\item \textbf{VLM judge:} $J(X,P)$  is the calibrated normalized Likert score of the plausibility estimate.
\item \textbf{Proxy judge ($\mathrm{VPTT_{score}}$):} for text-scale evaluation,
$J(X,P)$ is approximated by $\mathrm{VPTT_{score}}(p',P)$, shown in Sec.~4 of the
main paper to correlate with human judgments.
\end{itemize}

\begin{table*}[h!]
\centering
\scriptsize
\renewcommand{\arraystretch}{1.35}
\setlength{\tabcolsep}{6pt}
\begin{tabular}{lcccccc}
\toprule
\textbf{Metric / Model} &
\makecell[c]{\textbf{Flux DB-LoRA}\\ @50} &
\makecell[c]{\textbf{Viper}\\ @ 1000} &
\makecell[c]{\textbf{DrUM}\\ @1000} &
\makecell[c]{\textbf{Flux Kontext}\\ @1000} &
\makecell[c]{\textbf{GPT (VLM +}\\ \textbf{GPT-image-1} @100)}&
\makecell[c]{\textbf{GPT-image-1}\\ @100} \\
\midrule

\textbf{PS (Ours)} &
\makecell[c]{\textbf{0.867}\\{\tiny $\pm$ 0.173}} &

\makecell[c]{\textbf{0.678}\\{\tiny $\pm$ 0.269}} &

\makecell[c]{\textbf{0.841}\\{\tiny $\pm$ 0.139}} &
\makecell[c]{\textbf{0.839}\\{\tiny $\pm$ 0.184}} &
\makecell[c]{\textbf{0.858}\\{\tiny $\pm$ 0.180}} &
\makecell[c]{\textbf{0.974}\\{\tiny $\pm$ 0.019}} \\

\textbf{PS (Other)} &
\makecell[c]{0.656\\{\tiny $\pm$ 0.232}} &
\makecell[c]{0.545\\{\tiny $\pm$ 0.292}} &
\makecell[c]{0.757\\{\tiny $\pm$ 0.185}} &
\makecell[c]{0.541\\{\tiny $\pm$ 0.297}} &
\makecell[c]{0.832\\{\tiny $\pm$ 0.0.163}} &
\makecell[c]{0.966\\{\tiny $\pm$ 0.047}} \\

\textbf{PS Win \% (Ours)} &
\textbf{80.0} & \textbf{71.5} & \textbf{68.9} & \textbf{83.4} & \textbf{63.0} & 44.0 \\
\midrule

\textbf{VJ (Ours)} &
\makecell[c]{\textbf{3.17}\\{\tiny $\pm$ 0.70}} &
\makecell[c]{\textbf{2.88}\\{\tiny $\pm$ 1.13}} &
\makecell[c]{\textbf{3.41}\\{\tiny $\pm$ 0.57}} &
\makecell[c]{\textbf{3.11}\\{\tiny $\pm$ 0.73}} &
\makecell[c]{\textbf{3.28}\\{\tiny $\pm$ 0.75}} &
\makecell[c]{3.73 \\{\tiny $\pm$ 0.44}} \\

\textbf{VJ (Other)} &
\makecell[c]{1.99 \\{\tiny $\pm$ 0.67} } &
\makecell[c]{2.40 \\{\tiny $\pm$ 1.21} } &
\makecell[c]{2.89 \\{\tiny $\pm$ 0.61} } &
\makecell[c]{2.29\\{\tiny $\pm$ 0.98}} &
\makecell[c]{3.22\\{\tiny $\pm$ 0.71}} &
\makecell[c]{\textbf{3.86} \\{\tiny $\pm$ 0.35}} \\

\textbf{VJ Win \% (Ours)} &
\textbf{88.0 ( + 2\% ties)} & \textbf{61.8 (+15.5\% ties)} & \textbf{76.4 (+7.1\% ties)} & \textbf{76.4 (+ 7.9\% ties)} & \textbf{49.0 (+ 4\% ties)} & 33.0 ( + 12\% ties) \\

\bottomrule
\end{tabular}
\caption{Benchmark comparison on VIPER Proxy Score (PS) and VLM Judge Score (VJ). Mean and standard deviation appear on separate lines. Win \% reports the percentage of pairwise wins against the compared method for each metric.}
\label{tab:viper_flux_comparison}
\end{table*}

\begin{figure*}[t] 
    \centering \includegraphics[width=1.0\textwidth]{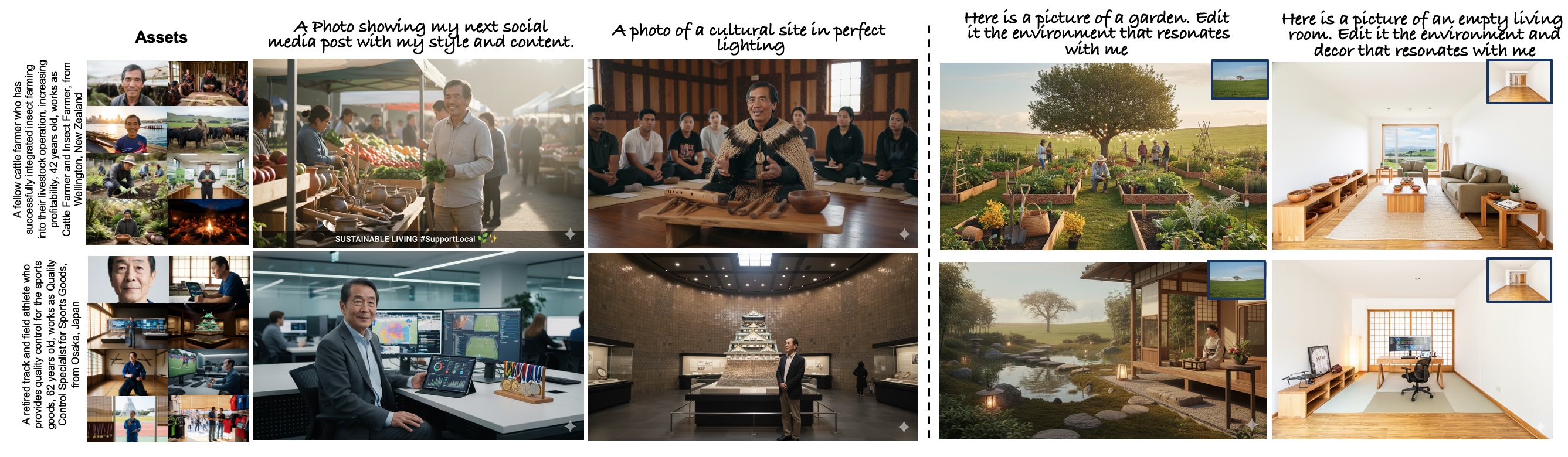}
    \caption{\textbf{Contextual Image Generation and Editing using VPTT-Bench.} Each row shows a distinct user profile: assets and style cues (left), personalized generations (social post, cultural site), and edits (garden, living room) guided by the same persona identity. All images are generated synthetically via our Visual Personalization RAG (VPRAG) by text, which retrieves persona-aligned cues. To show cross model personalization here the assets are generated by QWEN-image-model~\cite{yang2025qwen3technicalreport} and generations and edits by Nano-Banana~\cite{nano_banana} conditioned only on the first image.}
    \label{fig:main2}
\end{figure*}

\begin{figure*}[t] 
    \centering \includegraphics[width=1.0\textwidth]{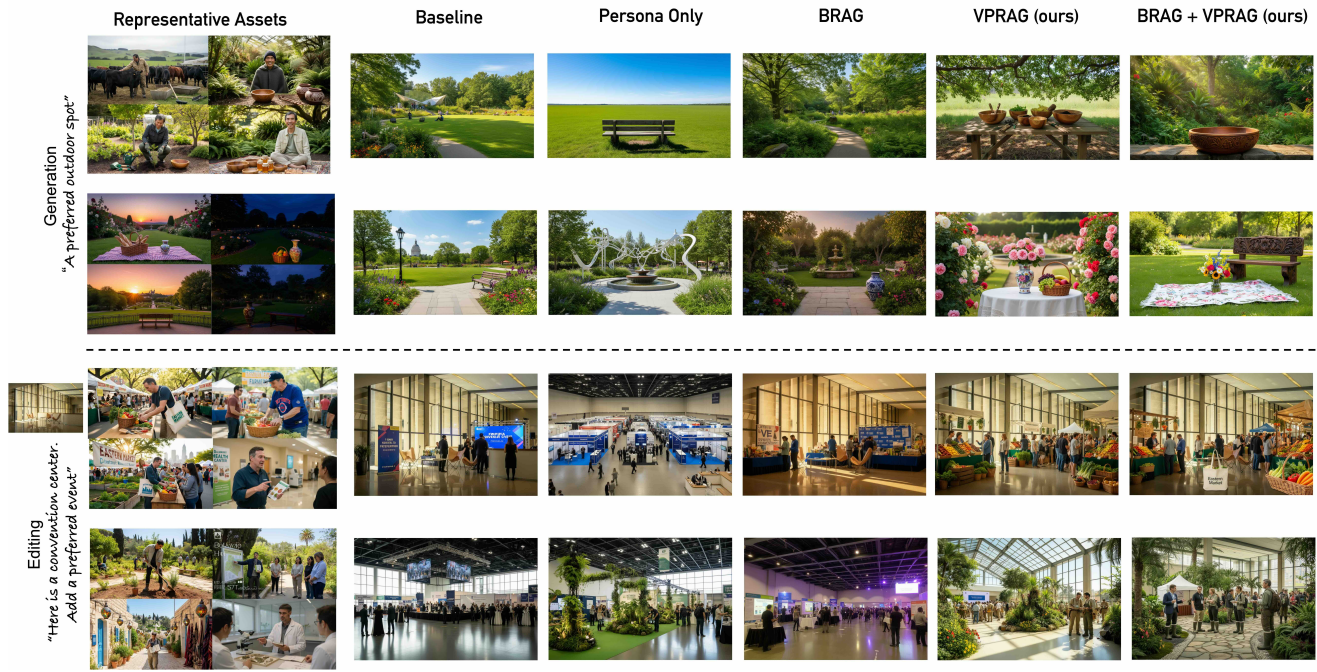}
    \caption{\textbf{Qualitative Comparison across Generation and Editing Tasks.}
Representative examples from the VPTT-Bench showing outputs from five methods: Baseline, Persona Only, BRAG, VPRAG (ours), and BRAG + VPRAG (ours).}
    \label{fig:more_comp}
\end{figure*}

\section{Limitations and Future Work}

\textbf{Synthetic--Real Gap.}  
Because VPTT-Bench relies on a single family of generator models for producing the synthetic personas, the benchmark inevitably inherits stylistic and cultural biases of those models. This “real-to-sim” gap limits how faithfully the benchmark captures the full diversity of real users. A promising direction is to construct future versions of VPTT-Bench using a heterogeneous ensemble of generators across organizations and training paradigms. We argue, however, that a unified v1 benchmark is an essential step: it moves the field from a data-zero regime to one where controlled, scalable, and privacy-safe personalization research becomes possible.

\textbf{Image-Only Scope.}  
While our retrieval and alignment mechanisms are modality-agnostic, this work focuses on image generation and image editing. Extending VPTT to videos, 3D assets, and multi-view content is a natural next step, requiring new alignment metrics and temporal-consistency modules.

\textbf{Scaling Beyond Individuals.}  
Our method currently models single-person personalization. Future work can expand VPTT toward \emph{societal personalization}: simulating communities, subcultures, and collective preference distributions. Such extensions could enable population-level evaluation, community-aware media generation, and product design aligned with specific cohorts.

\textbf{Enhanced Visual Grounding.}  
Persona assets are currently represented as rich textual “deferred renderings.” Future work may couple these with segmentation or detection models to retrieve visual elements directly from user images for opt-in users. This would enable stronger grounding on real visual evidence and more faithful scene composition.

\textbf{Structure Preservation.}  
Current generators, including those used in VPRAG, do not guarantee preservation of spatial layout during editing. Incorporating structure-aware diffusion models or control modules (e.g., depth/segmentation guidance) may improve fidelity for demanding edit tasks.

\textbf{Human-in-the-Loop Integration.}  
VPRAG can naturally operate as a ``visual copilot": retrieving user-specific cues, proposing edits, and letting the user refine preferences. Iterative preference learning, reinforcement from user feedback, and federated fine-tuning represent compelling next steps.

\textbf{Real-World Deployment.}  
Although we use synthetic personas for privacy reasons, the same dataset construction pipeline can be inverted to annotate and structure real user data in an opt-in or federated setting. This would enable applying the VPTT Framework directly on real personalization tasks while maintaining strong privacy guarantees.

\section{VPTT at scale}

\subsection{Analogy for Deferred Rendering in VPTT}

An analogy for our ``deferred rendering’’ process is an expert film critic. 
A critic invests considerable effort watching hundreds of movies (the expensive offline alignment) to internalize what makes a script succeed on screen. 
Once trained, the critic can read a new script and predict whether it would make a compelling film \emph{before} spending millions producing it.

Similarly, VPTT evaluates a candidate prompt against a persona’s visual identity in text form, first aligning Human/VLM judges/$\mathrm{VPTT_{score}}$ and using cheap and reliable $\mathrm{VPTT_{score}}$ \emph{before} generating any images. 
This enables early rejection of weak generations, reducing costly rendering and accelerating personalization at scale.

\subsection{Comparison to Visual Baselines}

Returning to our ``deferred rendering" analogy, VPTT allows us to evaluate a ``script’’ (a candidate prompt) before producing the ``film’’ (the final generated image).  
In this context, several existing personalization approaches can be interpreted as \textit{high-budget productions} that must render the entire film before knowing whether it works:

\begin{itemize}
    \item \textbf{Per-user finetuning methods} such as DreamBooth/LoRA~\cite{dblora} require retraining for each identity.
    \item \textbf{Preference-driven generation} systems such as VIPER and DrUM~\cite{ViPer, kim2025drawmindpersonalizedgeneration} rely on only matching the preferences from images and text.
    \item \textbf{Multimodal LLM pipelines} (e.g., OpenAI GPT-4o VLM + GPT-Image-1~\cite{openai2024gpt4ocard, gpt_image}, GPT-Image-1~\cite{gpt_image}, and Flux Kontext~\cite{labs2025flux}) operate as large black-box modules that jointly hallucinate alignment and appearance, but remain difficult to control or steer.
\end{itemize}

Because these methods must generate/input images to refine alignment, they cannot benefit from early rejection or even privacy safe benchmarks. They thus incur high latency, high cost, and weaker controllability. In contrast, our VPRAG approach evaluates alignment in text first using $\mathrm{VPTT_{score}}$, requiring no per-user training and no iterative image synthesis.

\subsubsection{Evaluation Metrics}

For evaluation, we assess generation quality using both automated metrics (VIPER proxy score~\cite{ViPer}, assigns higher scores to query images that share the preferred visual attributes) and human-aligned VLM judges (Aligned Gemini 2.0 Flash~\cite{flash_model}, see Sec 4 in the main paper), where judges compare baseline and personalized generations, i.e., using ``A preferred outdoor spot" or ``A Photo showing my next social media post with my style and content." and the personalized version per profile using VPRAG.

\subsubsection{Evaluation Protocol}

\paragraph{Flux DB-LoRA@50} We fine-tune FLUX.1-dev~\cite{flux_dev} using LoRA with rank 16 on attention layers, training for 1000 steps with the Prodigy optimizer and pivotal tuning on CLIP text encoder. Each user's LoRA is trained on 30 gallery images paired with a user-specific trigger word to learn personalized visual styles. Since training LoRAs is expensive, we train this baseline for 50 users to make this evaluation comprehensive.

\paragraph{VIPER@1000} We evaluate VIPER~\cite{ViPer}, a visual preference optimization baseline that personalizes SDXL~\cite{SDXL} by optimizing text-to-image alignment. For each user, VIPER retrieves the top-10 most similar gallery posts given the prompt and uses both the images and captions to compute visual preferences (positive and negative prompts) that guide generation toward user-preferred visual styles. The methods generate images from the test prompt ("A preferred outdoor spot") and the personalized one (VPRAG RAG) using SDXL-base-1.0~\cite{SDXL}. Evaluation is conducted on 1,000 users. Comparison is done against a $5\times2$ grid of reference images (10 gallery posts/ assets).

\paragraph{DrUM@1000} We evaluate DrUM~\cite{kim2025drawmindpersonalizedgeneration}, a baseline that personalizes image generation by conditioning on the user prompt history. For each user, DrUM retrieves the top-5 most similar captions from their gallery posts given the prompt and uses them as reference prompts with personalization strength $\alpha=0.5$. The methods generate images from the test prompt ("A preferred outdoor spot") and the personalized one (VPRAG RAG) using Stable Diffusion v1.5~\cite{sd_15}. Evaluation is conducted on 1,000 users, where comparison is done using both methods' outputs against a $5\times2$ grid of reference images (10 gallery posts/ assets).

\paragraph{FLUX Kontext @1000} We evaluate FLUX.1-Kontext-dev~\cite{Kontext} in-context learning capability by conditioning generation on a $5\times5$ grid of 25 reference images from each user's gallery. For each user, we generate two images: one with the base prompt alone and one with a persona-enhanced prompt i.e. VPRAG (both conditioned on the same reference grid), allowing us to assess whether in-context visual conditioning alone is sufficient for personalization. Here we evaluate 1000 users for comprehensive evaluation.

\paragraph{Large Multi-Modal Models} We compare VPRAG against OpenAI's GPT-Image-1~\cite{gpt_image} with two approaches: (1) \textbf{GPT (VLM + GPT-Image-1 @100)}~\cite{openai2024gpt4ocard, gpt_image} a visual analysis baseline where GPT-4o~\cite{openai2024gpt4ocard} analyzes a $5\times5$ grid of 25 user gallery images to extract visual style preferences (foreground, background, materials, objects, lighting, actions, environment, appearance) and generates a refined 3-phrase prompt that incorporates these elements alongside the base prompt ("A preferred outdoor spot"), and (2) \textbf{GPT-image-1 @100} visuals generated directly using $5\times5$ grid of 25 user gallery images by GPT-Image-1~\cite{gpt_image}. This evaluation used the unconstrained version of VPRAG with the prompt "A Photo showing my next social media post with my style and content." (Figure. 2 in the main paper). These are compared with the VPRAG augmented generations. 

The system prompt for the \textbf{GPT (VLM + Image Gen @100)} baseline is: 

\begin{tcolorbox}[colback=gray!10,colframe=black!60,sharp corners]
\ttfamily
You are an expert at analyzing visual styles and preferences from image collections.

\end{tcolorbox}
\begin{tcolorbox}[colback=gray!10,colframe=black!60,sharp corners]
\ttfamily

Analyze the provided images and create a detailed image generation prompt that will generate a new image matching both:

1. The requested text prompt
2. The visual elements from the reference images so the generated image looks like it's from the same gallery

Focus on visual elements like: foreground, background, materials, objects, lighting, actions, environment, appearance, etc.

Your prompt should be EXACTLY 3 short descriptive phrases separated by commas.

\end{tcolorbox}

The user prompt of the baseline is: 

\begin{tcolorbox}[colback=gray!10,colframe=black!60,sharp corners]
\ttfamily
Here are images from a user's profile. Analyze the visual style, color preferences, composition patterns, and aesthetic choices in these images.

Pay attention to: foreground elements, background, materials, objects, lighting, actions, environment, and overall composition.

Create an image generation prompt for: $"{base\_prompt}"$

\end{tcolorbox}

\begin{tcolorbox}[colback=gray!10,colframe=black!60,sharp corners]
\ttfamily

The prompt should incorporate visual elements from these images so the generated image feels like it's part of the same gallery.

IMPORTANT: Your response must be EXACTLY 3 short phrases separated by commas. 

\end{tcolorbox}

The prompt for the \textbf{GPT-image-1 @100} baseline is: 

\begin{tcolorbox}[colback=gray!10,colframe=black!60,sharp corners]
\ttfamily
Focus on visual elements like: foreground, background, materials, objects, lighting, actions, environment, appearance, etc. in the images in this grid. Generate an output image for $"{base\_prompt}"$ using these visual elements such that the resultant image also feels like it belongs to this profile.
\end{tcolorbox}

The prompt for the \textbf{GPT-image-1 @100} VPRAG is:

\begin{tcolorbox}[colback=gray!10,colframe=black!60,sharp corners]
\ttfamily
Using the visual style from these reference images, create: ${"personalized\_prompt"}$.
\end{tcolorbox}
\subsubsection{Results}

Table~\ref{tab:viper_flux_comparison} compares these baselines. Here, our method outperforms all the methods or has comparable performance with the large multi-model models. In Figure~\ref{fig:visual}, we show results on more nuanced and difficult examples, where VPRAG consistently emerges as an efficient and controllable personalization method, performing on par with or outperforming these substantially more expensive baselines.
While in-context learning (ICL) approaches such as GPT-4o~\cite{openai2024gpt4ocard} or GPT-Image-1~\cite{gpt_image} can condition generation on a set of reference images, they suffer from two fundamental limitations that our persona-based formulation directly addresses. 

\textbf{First, ICL is not controllable.} Without explicit structure, these models frequently copy or closely mimic individual gallery images rather than synthesizing novel content (see Figure~\ref{fig:copying}) from a coherent blend of visual attributes. In our evaluation, penalizing such copy-paste behavior results in a substantial performance drop for the ICL baseline (4.08 $\rightarrow$ 3.86; a 5.4\% decline), whereas our persona-enhanced method exhibits far greater robustness (3.83 $\rightarrow$ 3.73; only a 2.6\% decline). This indicates that our approach learns to aggregate and recompose visual patterns across multiple references instead of replicating isolated scenes. Moreover, ICL provides no principled mechanism for selecting which visual attributes to incorporate: all gallery images are treated uniformly, causing models to focus on salient but potentially irrelevant cues.

\textbf{Second, ICL is not scalable.} Performance degrades as gallery size increases due to context-window constraints and attention dilution, and inference cost grows linearly with the number of reference images ($\mathcal{O}(n)$). This makes evaluation impractical for settings requiring richer user histories or larger galleries.

\textbf{Third, ICL is not economically viable at scale.} At GPT-4o Vision pricing (approximately \$0.01 per processed image), a single personalized generation conditioned on a typical gallery (e.g. 25 images) incurs roughly \$0.25 in image-token cost alone. A user generating 100 personalized images would therefore cost about \$25; serving one million such users would exceed \$25\text{M}, excluding text-token fees and overhead. Additionally, these closed-source APIs impose rate limits and quota restrictions, rendering them unsuitable for high-volume, production-scale personalization workloads. 

As larger vision models continue to become more democratized and cost-efficient, VPTT will only become more practical and broadly applicable, enabling scaled evaluation across a wider range of personalization tasks.

\section{Additional Results}

Additional results as an extension to Figure 2 in the main paper are shown in Figure~\ref{fig:main2}. Additionally, more examples , also an extension to Figure 6 in the main paper are shown in Figure~\ref{fig:more_comp}.

\section{VLLM-Bench Construction (Detailed)}
\label{sec:persona_supp}

\subsection{Conceptual Basis: Deferred Rendering}
VLLM-Bench conceptualizes text generation as \emph{deferred rendering of visual identity}.  
Instead of pixels, each profile is expressed through language-level equivalents of visual cues i.e., objects, lighting, actions, background, foreground, materials, appearance, expressions, pose etc., that together represent how a concept would appear in visual media.  
This abstraction decouples personalization from rendering, enabling scalability, and privacy-preserving use.

\paragraph{Bidirectional Symmetry.}
The process is fully reversible:
\[ \small
\text{images} \rightarrow \text{caption} \rightarrow \text{visual elements} \rightarrow \text{preferences} \rightarrow \text{persona}.
\]
In forward mode, we generate structured text; in inverse mode, real user profiles can be captioned and converted into the same structure, enabling safe, text-only adaptation.

\subsection{Demographic Generation}

\subsubsection{Seed Initialization from PersonaHub}
We initialize 10K personas from \textbf{PersonaHub}~\cite{ge2025scalingsyntheticdatacreation}, which contains $\approx$200K curated human-authored persona descriptions.  
Each persona is derived using:
\begin{equation}
\text{seed\_index}_i = \text{hash}(i) \bmod 200{,}000,
\end{equation}
ensuring deterministic diversity across geography, age, and profession.

\subsubsection{Two-Stage Demographic Expansion}
\paragraph{Stage~1a: LLM-Based Demographics with Bias Mitigation.}
Given a seed $s$, we use Qwen2.5-72B-Instruct to infer country, city, ethnicity, and gender.  
When confidence in location grounding is low, hash-based diversity re-mapping adjusts sampling with region-specific override rates:
India/South~Asia~(70\%), United~States~(65\%), United~Kingdom~(60\%), and Canada~(50\%).  
This guarantees balanced representation across 9 ethnicity groups and 60+ authentic cities.

\paragraph{Stage~1b: Persona Completion.}
Demographic scaffolds are expanded into 20+ attributes, including occupation, education, interests (5–8 domain-specific), social-media tone, and lifestyle traits.  
Gender is inferred only when explicitly stated in the seed description; otherwise, it is marked as “unknown’’ to avoid introducing occupational or cultural stereotypes. While residual bias may still propagate through downstream image-generation models, the \textit{\textbf{final version of VPTT-Bench will include additional filtering and adjustments to mitigate such demographic biases}.}

\subsection{Visual Elements and Preference Generation}

Each persona contains a structured \textbf{visual vocabulary} with 15--20 entries per facet:
\begin{itemize}
    \item \textbf{Foreground:} subjects, actions, objects, body poses;
    \item \textbf{Background:} environments, landmarks, lighting, textures;
    \item \textbf{Atmospheric:} materials, color palette, mood, time of day.
\end{itemize}
At least 70\% of all elements reference culturally authentic motifs drawn from the persona’s region (e.g., Seoul Tower, Kashmiri gardens, or Venice canals).

We also generate 15--20 aesthetic and behavioral \textbf{preferences} (e.g., ``prefers warm lighting,'' ``posts minimalist compositions,'' ``documents festivals'') that act as latent conditioning factors. These are then used in feedback simulation part of the method to learn subjective preferences.

\subsection{Scenario and Caption Generation}
Each persona produces 30 posts in two phases:
\begin{enumerate}
    \item \textbf{Scenario Generation:} We sample 6--8 high-temperature ($\tau=0.9$) scenarios per batch with diversity constraints across content type (35\% activity, 25\% appreciation, 25\% shared content, 15\% selfie), temporal variety (day/night, seasonal), and social context (solo/group).
    \item \textbf{Caption Synthesis:} For each scenario, the LLM behaves as a vision-language model and produces a 150--250-word caption containing:
    (i) compositional details,  
    (ii) cultural context,  
    (iii) visible preferences, and  
    (iv) annotated facets (foreground, background, atmosphere).
\end{enumerate}
Each caption is encoded using the \texttt{text-embedding-3-small} model (1536\,D).  
Unused elements are pruned to ensure structural consistency.

\subsection{Parallelized Generation Pipeline}
The dataset is produced on an 8$\times$A100 GPU cluster with \textbf{vLLM-optimized Qwen2.5-72B}.  
Dynamic batching (10--200 profiles) yields 50--150 profiles/hour.  
Generation of 10K profiles (300K posts) completes in $\approx$66--200 hours. See Table.~\ref{tab:persona_supp_stats}.

\begin{table}[h]
\centering
\caption{\textbf{VPTT-Bench Generation Statistics.}}
\label{tab:persona_supp_stats}
\begin{tabular}{lc}
\toprule
\textbf{Metric} & \textbf{Value} \\
\midrule
Total personas & 10\,000 \\
Posts per persona & 30 \\
Total posts & 300\,000 \\
Mean caption length & 187.2 words \\
Mean visual elements/persona & 45.3 \\
Parallel throughput & 50--150 profiles/hr \\
\bottomrule
\end{tabular}
\end{table}

Profiles with fewer than 10 valid posts are excluded.  
All attributes, embeddings, and metadata are stored in JSONL format.

\subsection{Privacy, Scalability, and Extensibility}
Because all profiles are text-based, VPTT-Bench operates fully under deferred rendering, guaranteeing privacy and model-agnostic applicability.  
The dataset can be scaled to millions of profiles or augmented with real-world profiles through inverse captioning:
\[
\text{caption} \rightarrow \text{elements} \rightarrow \text{preferences} \rightarrow \text{persona}.
\]
This symmetric design ensures both the generative and analytical components of VPTT Framework can operate without any visual exposure, making VPTT-Bench a reusable personalization benchmark.

\section{Visual Assets Generation}
\label{supp:image_generation}

\subsection{Mathematical Face Diversity System}
To ensure controlled, globally diverse identity synthesis, we implement a deterministic facial attribute generator producing $97.2$M unique combinations. These are then added to the demographics description to first generate a user persona image and then conditioned on this image to generate 30 assets.

\paragraph{Attribute Space.}
Ten facial attributes with 4–6 discrete options each are defined: 
face shape, eye shape, eye size, nose type, jawline, cheekbones, lip shape, eyebrow type, face length, and chin shape.  
For each user ID and age group, we compute:
\[
\text{seed} = \text{hash}(\text{user\_id}, \lfloor \tfrac{\text{age}}{10} \rfloor) \bmod 2^{32},
\]
and draw attributes $F = \{a_1,\dots,a_{10}\}$ from the option sets $\{O_i\}$.  
Modifiers such as age-adapted details (e.g., ``bright eyes'' vs. ``wisdom lines''), expression labels, and photo styles are applied to achieve additional realism.

\paragraph{Combinatorial Diversity.}
\[
N = \prod_{i=1}^{10}|O_i| \times |A| \times |E| \times |S| \approx 9.72{\times}10^{7},
\]
where $A$ denotes age modifiers, $E$ expression states, and $S$ photo styles.  
This formulation ensures reproducible sampling and balanced variation across users.

\subsection{Two-Phase Image Generation Pipeline}

\paragraph{Phase 1: Persona Base Generation (Text-to-Image).}
Each persona’s base portrait is synthesized using Qwen-Image 2509~\cite{yang2025qwen3technicalreport, 2509} diffusion models.  
Prompts combine demographics and generated facial attributes:
\begin{quote}
\small
``photo of a person, \{gender\}, \{race/ethnicity\}, \{age\} years old, works as \{occupation\}, from \{city, country\}, 
\{oval face shape\}, \{almond eyes\}, \{high cheekbones\}, \{full lips\}, 
professional portrait, confident expression, natural lighting.''
\end{quote}
\noindent
Generation parameters:
\begin{itemize}
    \item Model: Qwen-Image or Qwen-Image-Edit (vanilla mode)
    \item Resolution: 1344×768
    \item Steps: 50, CFG=0.0, Seed=Deterministic per user
    \item No negative prompts (maximizes diversity)
\end{itemize}

\paragraph{Phase 2: Post-Specific Editing (Image-to-Image).}
Each persona’s 30 textual posts is rendered by Qwen-Image-Edit-2509.  
Prompts differ by post type:
\begin{itemize}
    \item \textbf{Activity / Selfie / Shared Content (70\%):}
    \begin{quote}
    ``\{caption\}, wearing \{clothing\}, with \{expression\}, \{pose\}.''
    \end{quote}
    \item \textbf{Appreciation Posts (30\%):}
    \begin{quote}
    Prompt: ``\{scene\_description\}''  
    Negative: ``person, people, human, face, body, portrait''
    \end{quote}
\end{itemize}

\paragraph{Configurations.}
\begin{itemize}
    \item \textbf{Standard Mode:} 40 steps, CFG=4.0, $\sim$15–20\,s/image
    \item \textbf{Lightning LoRA Mode:} 4–8 steps, CFG=4.5, $\sim$3–5\,s/image (4× faster)
\end{itemize}
Pronouns are replaced with ``this person'' to ensure gender neutrality.

\subsection{Parallel Multi-GPU System}
An 8$\times$A100 cluster executes both phases in parallel.  
Models are cached per GPU; tasks are dynamically queued via thread-safe managers to maintain 100\% utilization.  
Phase 1 (base portraits) and Phase 2 (post edits) can run independently or sequentially (Table.~\ref{tab:supp_image_perf}).

\begin{table}[h]
\small
\centering
\caption{Synthetic Image Generation Performance Metrics.}
\label{tab:supp_image_perf}
\begin{tabular}{lc}
\toprule
\textbf{Metric} & \textbf{Value} \\
\midrule
Throughput (standard) & 50--80 images/hour/GPU \\
Throughput (Lightning) & 180--240 images/hour/GPU \\
Memory footprint & $<$24 GB/GPU (bfloat16 precision) \\
\bottomrule
\end{tabular}
\end{table}

\section{VPTT-Bench Stats}

To illustrate the diversity of the benchmark, Figure~\ref{fig:diversity1} presents the distribution of ethnicities and countries of origin. Despite the modest number of samples, the population is highly diverse. Similarly, Figure~\ref{fig:diversity2} reports the age distribution of the benchmark and the interests of the first 1,000 users, grouped by ethnicity. At a larger scale, Figure~\ref{fig:diversity3} visualizes the averaged caption embeddings of 10K users, highlighting diversity across age groups and visual attributes.

\section{VPRAG Algorithm}

To formally define the steps used by our VPRAG method, Algorithm~\ref{alg:visual_rag_concise} shows a compact form of the retrival engine. 

\begin{algorithm}[t]
\caption{VPRAG with Optional Feedback Re-ranking}
\label{alg:visual_rag_concise}
\begin{algorithmic}[1]
\STATE \textbf{Inputs:} query $p$, persona memory $\mathcal{M}=\{(E_i, c_i)\}_{i=1}^{N}$, categories $\mathcal{C}$, budgets $\{Q^{(k)}\}$, temperature $\tau$, category embeddings $\{\mathbf{u}_k\}$
\STATE \textbf{Embedders:} post: $\text{Embed}_{\text{OpenAI}}$, element: $\text{Embed}_{\text{MiniLM}}$
\STATE \textbf{Outputs:} re-prompt $p'$, (optional) re-ranked $p'^*$

\STATE \textbf{Post-level retrieval:}
\STATE $\mathbf{q}\!\leftarrow\!\frac{\text{Embed}_{\text{OpenAI}}(p)}{\|\cdot\|_2}$;\quad $\mathbf{v}_i\!\leftarrow\!\frac{\text{Embed}_{\text{OpenAI}}(c_i)}{\|\cdot\|_2}$
\STATE $w_i \leftarrow \frac{\exp(\mathbf{q}^\top \mathbf{v}_i/\tau)}{\sum_j \exp(\mathbf{q}^\top \mathbf{v}_j/\tau)}$;\quad $H \leftarrow -\sum_i w_i\log w_i$;\quad $n_{\text{eff}}\!\leftarrow\!\exp(H)$
\STATE $Q \leftarrow \sum_{k} Q^{(k)}$;\quad $K \leftarrow \min\big(\lfloor n_{\text{eff}} \rfloor,\, 2Q\big)$
\STATE $\mathcal{I} \leftarrow \text{TopKIndices}(w, K)$

\STATE \textbf{Category priorities \& quotas:}
\STATE $\text{priority}_k \leftarrow \mathbf{q}^\top \mathbf{u}_k$;\quad $\mathcal{C}_{\text{sorted}}\!\leftarrow\!\text{SortBy}(\text{priority}_k)$
\FOR{$k \in \mathcal{C}_{\text{sorted}}$}
  \STATE $c_i^{(k)} \leftarrow |E_i^{(k)}|,\; i\!\in\!\mathcal{I}$;\quad
  $q_i^{(k)} \leftarrow \Big\lfloor \frac{w_i c_i^{(k)}}{\sum_{j\in\mathcal{I}} w_j c_j^{(k)}} Q^{(k)} \Big\rfloor$
\ENDFOR

\STATE \textbf{Element ranking (atomic):}
\STATE $\mathbf{q}_{\text{elm}} \leftarrow \frac{\text{Embed}_{\text{MiniLM}}(p)}{\|\cdot\|_2}$;\quad $\mathcal{E}_p \leftarrow \emptyset$
\FOR{$k \in \mathcal{C}_{\text{sorted}}$} \FOR{$i \in \mathcal{I}$}
  \STATE \textbf{if} $q_i^{(k)}{=}0$ \textbf{then} continue
  \STATE \small $\mathcal{S}_{i,k}\!\leftarrow\!\text{TopK}\big(E_i^{(k)},\, q_i^{(k)};\; \cos(\text{Embed}_{\text{MiniLM}}(\cdot),\mathbf{q}_{\text{elm}})\big)$
  \STATE $\mathcal{E}_p \leftarrow \mathcal{E}_p \cup \mathcal{S}_{i,k}$
\ENDFOR \ENDFOR

\STATE \textbf{Compose:}\; $p' \leftarrow f_{\text{compose}}(p, \mathcal{E}_p)$ \COMMENT{or $f_{\text{compose}}(p,\mathcal{S}_p,\mathcal{E}_p)$ if $\mathcal{S}_p$ is precomputed}

\STATE \textbf{(Optional) feedback re-ranking:}
\STATE Train small $f_\theta$ on few profiles: $(p',\mathcal{P}) \mapsto s_{\text{VLM}}\!\in\![0,1]$
\STATE At inference, sample $\{p'_m\}_{m=1}^M$ and pick $p'^* \!=\! \arg\max_m f_\theta(\text{Embed}(p'_m),\text{Embed}(\mathcal{P}))$

\STATE \textbf{return} $p'$ (or $p'^*$)
\end{algorithmic}
\end{algorithm}

\section{Real-World Examples}

Only for demonstration purposes, we include a small set of real-world example images that illustrate the types of visual inputs supported by our method. These images are not part of the training set, evaluation benchmarks, or any quantitative analysis; they are shown solely to help readers qualitatively understand the range of scenarios in which the system operates.

The human figures (Figures.~\ref{fig:user1} and \ref{fig:user2}) shown in the sample images are non-author volunteers who provided informed consent for their anonymized photos to be used for illustration. All faces and identifying features (e.g., facial attributes, backgrounds revealing location) have been fully obscured to preserve privacy. These individuals have no relationship to the authorship of the paper, and their inclusion does not reveal author identity in any way.

These examples highlight the diversity of environments, poses, and visual conditions encountered in typical user-generated content, and demonstrate how the proposed system generalizes across varied real-world scenes.

\section{Expanded Tables.}

Tables~\ref{tab:method_comparison_stacked}, 
\ref{tab:method_comparison_stacked_social}, \ref{tab:method_comparison_stacked_living_room}, \ref{tab:method_comparison_stacked_garden}, shows the expanded versions of the Table 3 in the main paper. Here we report both $\mathrm{VPTT_{score}\text{-}c}$ and $\mathrm{VPTT_{score}}$ scores showing the results are consistent with the experiments in the main paper. Cohen's d in these table are computed against the Baseline.

\section{User Study Protocol}

To measure how well generated images align with an individual's visual style, we conducted a human evaluation following the VPTT. Each task presented annotators with a 10–image gallery representing a user's typical aesthetics, environments, lighting preferences, clothing patterns, and recurring visual motifs. Alongside the gallery, annotators viewed a 2$\times$2 grid of generated images (Methods A–D). Participants rated each generated image independently using a slider ranging from 0 to 5, guided strictly by visual similarity to the user's gallery rather than image quality, personal preference, or cross-method comparison.

This setup allowed us to isolate whether a generated sample \emph{belonged to the same visual world} as the user's posts. Annotators were trained to focus on concrete signals such as objects, materials, environments, appearance patterns, lighting, and cultural or stylistic markers. By collecting similarity judgments across thousands of examples, we obtained a fine-grained human signal for the plausibility and consistency of personalization across diverse prompts and visual domains. Here is a concise form of the instructions: 

\begin{tcolorbox}[colback=gray!10,colframe=black!60,sharp corners]
\ttfamily
\textbf{Annotation Instructions} \\[2pt]
Rate each generated image from 0 to 5 based on: \\[2pt]
-- Objects \& Materials: Are key objects or textures similar to those in the gallery? \\
-- Environments \& Settings: Are backgrounds or locations consistent with the user's style? \\
-- Appearance Patterns: Clothing style, color palette, accessories, poses? \\
-- Lighting \& Atmosphere: Similar mood, time of day, natural/white lighting? \\
-- Cultural / Style Markers: Recurring themes, sports references, regional dress, etc. \\[4pt]

\textbf{Do NOT:} rate based on image quality, personal preference, comparison across methods, or prompt correctness. \\[4pt]

\textbf{Score Guide:} \\
5 = Excellent match (fits naturally in user's gallery) \\
4--4.5 = Good match \\
3--3.5 = Moderate similarity \\
2--2.5 = Weak similarity \\
1--1.5 = Minimal similarity \\
0--0.5 = No similarity at all

\end{tcolorbox}

\subsection{VLM Judge for Automatic Persona Evaluation}

To complement the human user study, we use a vision–language model (VLM) as an automatic judge that approximates the same visual-similarity protocol. For each user in the \textit{generation} split, we first construct a \emph{profile canvas} by tiling up to 10 of their posts into a 5$\times$2 grid, with each post numbered. We then construct a \emph{methods canvas} by arranging the five generated images from different methods horizontally and assigning them blind labels A--E via a user-specific but deterministic shuffle. The VLM receives the baseline textutal generation prompt, the profile canvas, and the methods canvas as inputs, and is asked to score each of A--E independently on a 0--5 scale based purely on visual similarity to the gallery, mirroring the human instructions.

For the \textit{editing} split, the setup is identical except that we additionally provide the original input image that was edited. The same VLM judge prompt structure is used, but the user message explicitly refers to an editing task and includes the editing prompt. In both cases, we query either GPT-4o Vision or Gemini-2.5-Pro (to remove the model bias for the generations by 4o-mini or Gemini-2.5-Pro) with a fixed system instruction and a task-specific user instruction. The model returns natural-language lines that we parse into per-method scores in $[0,5]$ (with 0.5 increments) and short explanations.

\begin{tcolorbox}[colback=gray!10,colframe=black!60,sharp corners]
\ttfamily
\textbf{System prompt (VLM judge):} \\[4pt]
You are an expert visual judge evaluating AI-generated images for visual similarity to a user's persona.

Evaluate how well each generated image captures the VISUAL ELEMENTS from the persona's posts.

**EVALUATION CRITERIA (Visual Similarity):**

1. **Objects \& Materials**: Same objects, materials, textures visible in posts?
2. **Environment \& Setting**: Similar locations, backgrounds, environments?
3. **Appearance Patterns**: Similar clothing styles, colors, expressions?
4. **Lighting \& Atmosphere**: Similar lighting, mood, atmosphere?
5. **Colors \& Composition**: Similar color palettes and visual composition?
6. **Cultural/Style Markers**: Similar cultural elements, aesthetic style?

**SCORING (0-5 scale, use 0.5 increments):**
- **5.0**: Excellent visual similarity - Captures most key visual elements from posts
- **4.0-4.5**: Good visual similarity - Several key visual elements present
- **3.0-3.5**: Moderate visual similarity - Some visual elements recognizable
- **2.0-2.5**: Weak visual similarity - Few visual elements match
- **1.0-1.5**: Minimal visual similarity - Barely any visual connection
- **0.0-0.5**: No visual similarity - Completely different visual style

\end{tcolorbox}

\begin{tcolorbox}[colback=gray!10,colframe=black!60,sharp corners]
\ttfamily

**IMPORTANT**: Focus on VISUAL similarity, not just conceptual alignment.
\end{tcolorbox}

\begin{tcolorbox}[colback=gray!10,colframe=black!60,sharp corners]
\ttfamily
\textbf{User prompt (generation tasks):} \\[4pt]
TASK: Evaluate these 5 generated images for visual similarity to the persona's posts

**GENERATION PROMPT:** "\{base\_prompt\}"

**IMAGE 1 (Reference - Profile Context):**
Grid of selected posts from the persona's gallery (numbered).

**IMAGE 2 (Generated Images to Evaluate):**
5 generated images labeled A through E (left to right).
Each was generated using a different approach (you don't know which approach was used for which image).

**YOUR TASK:**
For EACH image (A, B, C, D, E), score 0-5 based on:
- How similar are the VISUAL ELEMENTS (objects, environment, appearance, lighting, colors)?

- Do the generated images look like they could belong to the same persona's gallery?
- Are there recognizable visual patterns from the posts?

Respond in this EXACT format (one line per image):
A: Score=X.X - [1-2 sentence explanation of why this score, focusing on specific visual elements]
B: Score=Y.Y - [1-2 sentence explanation of why this score, focusing on specific visual elements]
C: Score=Z.Z - [1-2 sentence explanation of why this score, focusing on specific visual elements]
D: Score=W.W - [1-2 sentence explanation of why this score, focusing on specific visual elements]
E: Score=V.V - [1-2 sentence explanation of why this score, focusing on specific visual elements]

\end{tcolorbox}

\begin{tcolorbox}[colback=gray!10,colframe=black!60,sharp corners]
\ttfamily
\textbf{User prompt (editing tasks):} \\[4pt]
TASK: Evaluate these 5 edited images for visual similarity to the persona's posts

**EDITING PROMPT:** "\{base\_prompt\}"

**IMAGE 1 (Input Image):**
The original input image that was edited.

**IMAGE 2 (Reference - Profile Context):**
Grid of selected posts from the persona's gallery (numbered).

**IMAGE 3 (Edited Images to Evaluate):**
5 edited images labeled A through E (left to right).
Each was edited using a different approach (you don't know which approach was used for which image).

**YOUR TASK:**
For EACH image (A, B, C, D, E), score 0-5 based on:
- How similar are the VISUAL ELEMENTS (objects, environment, appearance, lighting, colors)?
- Do the edited images look like they could belong to the same persona's gallery?
- Are there recognizable visual patterns from the posts?

Respond in this EXACT format (one line per image):
A: Score=X.X - [1-2 sentence explanation of why this score, focusing on specific visual elements]
B: Score=Y.Y - [1-2 sentence explanation of why this score, focusing on specific visual elements]
C: Score=Z.Z - [1-2 sentence explanation of why this score, focusing on specific visual elements]
D: Score=W.W - [1-2 sentence explanation of why this score, focusing on specific visual elements]
E: Score=V.V - [1-2 sentence explanation of why this score, focusing on specific visual elements]

\end{tcolorbox}

\section{Implementation Details}

For Table 3 in the main paper, we use Qwen2.5-7B-Instruct via vLLM for text generation ($T=0.1$, top-p $=0.9$, max tokens $=256$, seed $=42$). For GPT-4o-mini, we use $T=0.1$, seed $=42$.  We use Gemini-2.5-Pro with $T=0.7$, top-p $=0.95$. This temperature is chosen as we noticed that lower temperatures tend to truncate the text.  Our soft assignment mechanism computes post-level attention weights via softmax with temperature $\tau=0.1$. 

For the VLM judges in the main paper, we use Gemini-2.5-Pro with temperature $T=0.0$, top-p $p=0.95$, and maximum output tokens of 5000. Another variant is the GPT-4o Vision with temperature $T=0.0$ , and seed = $42$.

For the feedback network, we train a lightweight cross-attention transformer to predict user-prompt alignment scores. The model takes text-embedding-3-small (1536-dim) embeddings of user profiles and prompts as input, projecting them to a 128-dimensional hidden space. Cross-attention with 4 heads allows the profile representation to attend to prompt features, followed by a feed-forward network (128 → 256 → 128) with residual connections and layer normalization. The final prediction head (256 → 128 → 64 → 1) outputs scores in [0,1] via sigmoid activation. We train with AdamW (lr=0.001, weight decay=0.05) using MSE loss, with dropout=0.2 for regularization and early stopping (patience=10).

\begin{table*}[h]
\centering
\caption{Method Comparison Across LLMs (Cultural Site Prompt)}
\label{tab:method_comparison_stacked}
\resizebox{\textwidth}{!}{
\begin{tabular}{l|l|cccc|ccc|ccc}
\hline
\multirow{3}{*}{\textbf{LLM}} & \multirow{3}{*}{\textbf{Method}} & \multicolumn{4}{c|}{\textbf{Individual Metrics}} & \multicolumn{3}{c|}{\textbf{$\mathrm{VPTT_{score}\text{-}c}$} (Uniform Weights)} & \multicolumn{3}{c}{\textbf{$\mathrm{VPTT_{score}}$} (Novelty Adjusted)} \\
& & \textbf{PA} & \textbf{GS} & \textbf{CP} & \textbf{NV} & \textbf{Score} & \textbf{Win\%} & \textbf{d} & \textbf{Score} & \textbf{Win\%} & \textbf{d} \\
& & & & & & & & \textit{(vs base)} & & & \textit{(vs base)} \\
\hline
\multirow{5}{*}{4o-mini} & Baseline & 0.150 & 0.375 & 0.589 & -- & 0.371 & 0.0\% & -- & 0.319 & 0.0\% & -- \\
 & Persona Only & 0.364 & 0.429 & 0.630 & -- & 0.474 & 0.1\% & 4.18 & 0.391 & 0.0\% & 3.62 \\
 & BRAG & 0.316 & \underline{0.597} & \underline{0.674} & \underline{0.858} & 0.529 & 8.7\% & 4.40 & 0.616 & 11.5\% & 10.94 \\
 & VPRAG (Ours) & \underline{0.401} & 0.591 & 0.660 & \textbf{0.900} & \underline{0.551} & \underline{17.9}\% & \underline{5.76} & \underline{0.635} & \underline{31.8}\% & \textbf{13.19} \\
 & Comb. VPRAG+BRAG (Ours) & \textbf{0.419} & \textbf{0.641} & \textbf{0.686} & 0.821 & \textbf{0.582} & \textbf{73.3}\% & \textbf{5.95} & \textbf{0.646} & \textbf{56.8}\% & \underline{12.73} \\
\hline
\multirow{5}{*}{Qwen} & Baseline & 0.150 & 0.375 & 0.589 & -- & 0.371 & 0.0\% & -- & 0.319 & 0.0\% & -- \\
 & Persona Only & 0.325 & 0.417 & 0.627 & -- & 0.456 & 0.1\% & 3.45 & 0.378 & 0.0\% & 3.00 \\
 & BRAG & \underline{0.395} & \textbf{0.670} & \textbf{0.683} & 0.441 & \textbf{0.583} & \textbf{48.6}\% & \underline{5.38} & 0.573 & 14.5\% & 6.34 \\
 & VPRAG (Ours) & 0.378 & 0.587 & 0.656 & \textbf{0.863} & 0.540 & 11.5\% & 5.35 & \textbf{0.621} & \textbf{62.0}\% & \textbf{12.24} \\
 & Comb. VPRAG+BRAG (Ours) & \textbf{0.414} & \underline{0.649} & \underline{0.678} & \underline{0.544} & \underline{0.580} & \underline{39.9}\% & \textbf{5.43} & \underline{0.590} & \underline{23.5}\% & \underline{6.93} \\
\hline
\multirow{5}{*}{Gemini} & Baseline & 0.150 & 0.375 & 0.589 & -- & 0.371 & 0.0\% & -- & 0.319 & 0.0\% & -- \\
 & Persona Only & 0.278 & 0.407 & 0.614 & -- & 0.433 & 0.1\% & 2.23 & 0.362 & 0.0\% & 2.00 \\
 & BRAG & 0.286 & \underline{0.606} & 0.647 & \underline{0.775} & 0.513 & 18.2\% & 3.12 & 0.588 & 11.6\% & 7.82 \\
 & VPRAG (Ours) & \textbf{0.359} & 0.597 & \underline{0.656} & \textbf{0.893} & \underline{0.537} & \underline{31.9}\% & \underline{4.69} & \textbf{0.626} & \textbf{58.0}\% & \textbf{11.30} \\
 & Comb. VPRAG+BRAG (Ours) & \underline{0.349} & \textbf{0.635} & \textbf{0.667} & 0.768 & \textbf{0.550} & \textbf{49.8}\% & \textbf{4.70} & \underline{0.614} & \underline{30.4}\% & \underline{9.46} \\
\hline
\end{tabular}
}
\end{table*}
\begin{table*}[h]
\centering
\caption{Method Comparison Across LLMs (Social Media Post Prompt)}
\label{tab:method_comparison_stacked_social}
\resizebox{\textwidth}{!}{
\begin{tabular}{l|l|cccc|ccc|ccc}
\hline
\multirow{3}{*}{\textbf{LLM}} & \multirow{3}{*}{\textbf{Method}} & \multicolumn{4}{c|}{\textbf{Individual Metrics}} & \multicolumn{3}{c|}{\textbf{$\mathrm{VPTT_{score}\text{-}c}$}(Uniform Weights)} & \multicolumn{3}{c}{\textbf{$\mathrm{VPTT_{score}}$}(Novelty Adjusted)} \\
& & \textbf{PA} & \textbf{GS} & \textbf{CP} & \textbf{NV} & \textbf{Score} & \textbf{Win\%} & \textbf{d} & \textbf{Score} & \textbf{Win\%} & \textbf{d} \\
& & & & & & & & \textit{(vs base)} & & & \textit{(vs base)} \\
\hline
\multirow{5}{*}{4o-mini} & Baseline & 0.174 & 0.322 & 0.602 & -- & 0.366 & 0.0\% & -- & 0.312 & 0.0\% & -- \\
 & Persona Only & 0.428 & 0.437 & 0.659 & -- & 0.508 & 0.9\% & 5.85 & 0.414 & 0.0\% & 5.34 \\
 & BRAG & 0.434 & \textbf{0.583} & \textbf{0.707} & 0.828 & \underline{0.574} & \underline{35.3}\% & 5.45 & 0.639 & 29.1\% & 11.65 \\
 & VPRAG (Ours) & \textbf{0.451} & 0.566 & 0.685 & \textbf{0.899} & 0.567 & 27.2\% & \textbf{6.17} & \textbf{0.645} & \textbf{40.5}\% & \textbf{13.30} \\
 & Comb. VPRAG+BRAG (Ours) & \underline{0.448} & \underline{0.581} & \underline{0.703} & \underline{0.837} & \textbf{0.578} & \textbf{36.6}\% & \underline{6.09} & \underline{0.643} & \underline{30.4}\% & \underline{12.64} \\
\hline
\multirow{5}{*}{Qwen} & Baseline & 0.174 & 0.322 & 0.602 & -- & 0.366 & 0.0\% & -- & 0.312 & 0.0\% & -- \\
 & Persona Only & 0.384 & 0.422 & 0.656 & -- & 0.488 & 0.0\% & 4.66 & 0.400 & 0.0\% & 4.33 \\
 & BRAG & \textbf{0.516} & \textbf{0.697} & \textbf{0.707} & 0.323 & \textbf{0.640} & \textbf{81.6}\% & \textbf{8.50} & 0.589 & 14.7\% & 7.23 \\
 & VPRAG (Ours)& 0.448 & 0.583 & 0.685 & \textbf{0.854} & 0.572 & 6.1\% & \underline{6.03} & \textbf{0.641} & \textbf{59.3}\% & \textbf{12.52} \\
 & Comb. VPRAG+BRAG (Ours) & \underline{0.456} & \underline{0.600} & \underline{0.702} & \underline{0.663} & \underline{0.586} & \underline{12.3}\% & 5.68 & \underline{0.614} & \underline{26.0}\% & \underline{8.54} \\
\hline
\multirow{5}{*}{Gemini} & Baseline & 0.174 & 0.322 & 0.602 & -- & 0.366 & 0.0\% & -- & 0.312 & 0.0\% & -- \\
 & Persona Only & 0.371 & 0.424 & 0.647 & -- & 0.481 & 0.0\% & 4.43 & 0.396 & 0.0\% & 4.11 \\
 & BRAG & \textbf{0.497} & \textbf{0.638} & \underline{0.694} & 0.724 & \textbf{0.610} & \textbf{60.7}\% & \textbf{6.96} & \underline{0.644} & \underline{39.0}\% & \underline{11.78} \\
 & VPRAG (Ours) & 0.396 & 0.544 & 0.674 & \textbf{0.894} & 0.538 & 3.7\% & 4.91 & 0.623 & 13.9\% & 11.66 \\
 & Comb. VPRAG+BRAG (Ours) & \underline{0.477} & \underline{0.610} & \textbf{0.699} & \underline{0.808} & \underline{0.595} & \underline{35.5}\% & \underline{6.25} & \textbf{0.650} & \textbf{47.1}\% & \textbf{12.10} \\
\hline
\end{tabular}
}
\end{table*}
\begin{table*}[h]
\centering
\caption{Method Comparison Across LLMs (Empty Living Room Prompt)}
\label{tab:method_comparison_stacked_living_room}
\resizebox{\textwidth}{!}{
\begin{tabular}{l|l|cccc|ccc|ccc}
\hline
\multirow{3}{*}{\textbf{LLM}} & \multirow{3}{*}{\textbf{Method}} & \multicolumn{4}{c|}{\textbf{Individual Metrics}} & \multicolumn{3}{c|}{\textbf{$\mathrm{VPTT_{score}\text{-}c}$} (Uniform Weights)} & \multicolumn{3}{c}{\textbf{$\mathrm{VPTT_{score}}$} (Novelty Adjusted)} \\
& & \textbf{PA} & \textbf{GS} & \textbf{CP} & \textbf{NV} & \textbf{Score} & \textbf{Win\%} & \textbf{d} & \textbf{Score} & \textbf{Win\%} & \textbf{d} \\
& & & & & & & & \textit{(vs base)} & & & \textit{(vs base)} \\
\hline
\multirow{5}{*}{4o-mini} & Baseline & 0.115 & 0.350 & 0.571 & -- & 0.346 & 0.0\% & -- & 0.299 & 0.0\% & -- \\
 & Persona Only & \underline{0.356} & 0.418 & \underline{0.635} & -- & 0.470 & 4.0\% & \underline{5.15} & 0.387 & 0.0\% & 4.56 \\
 & BRAG & 0.264 & 0.533 & 0.617 & \textbf{0.928} & 0.472 & 5.1\% & 3.80 & 0.584 & 8.9\% & 11.35 \\
 & VPRAG (Ours) & \textbf{0.379} & \textbf{0.560} & \textbf{0.654} & 0.908 & \textbf{0.531} & \textbf{81.4}\% & \textbf{5.68} & \textbf{0.622} & \textbf{77.7}\% & \textbf{13.07} \\
 & Comb. VPRAG+BRAG (Ours) & 0.306 & \underline{0.540} & 0.630 & \underline{0.924} & \underline{0.492} & \underline{9.4}\% & 4.51 & \underline{0.597} & \underline{13.5}\% & \underline{12.24} \\
\hline
\multirow{5}{*}{Qwen} & Baseline & 0.115 & 0.350 & 0.571 & -- & 0.346 & 0.0\% & -- & 0.299 & 0.0\% & -- \\
 & Persona Only & \underline{0.357} & 0.396 & 0.630 & -- & 0.461 & 0.3\% & \underline{5.14} & 0.379 & 0.0\% & 4.41 \\
 & BRAG & 0.339 & \underline{0.547} & \textbf{0.658} & 0.819 & 0.514 & \underline{18.9}\% & 4.56 & 0.593 & \underline{13.8}\% & 10.67 \\
 & VPRAG (Ours)& \textbf{0.416} & \textbf{0.583} & 0.657 & \textbf{0.873} & \textbf{0.552} & \textbf{66.7}\% & \textbf{6.45} & \textbf{0.630} & \textbf{76.6}\% & \textbf{13.54} \\
 & Comb. VPRAG+BRAG (Ours) & 0.348 & 0.540 & \underline{0.657} & \underline{0.828} & \underline{0.515} & 14.1\% & 4.31 & \underline{0.594} & 9.7\% & \underline{10.68} \\
\hline
\multirow{5}{*}{Gemini} & Baseline & 0.115 & 0.350 & 0.571 & -- & 0.346 & 0.0\% & -- & 0.299 & 0.0\% & -- \\
 & Persona Only & \underline{0.333} & 0.400 & 0.630 & -- & 0.454 & 3.6\% & 4.29 & 0.375 & 0.0\% & 3.80 \\
 & BRAG & 0.292 & \underline{0.529} & 0.619 & \underline{0.915} & 0.480 & 15.8\% & 3.92 & 0.586 & 16.2\% & 11.07 \\
 & VPRAG (Ours) & 0.322 & 0.526 & \underline{0.639} & \textbf{0.937} & \underline{0.496} & \underline{30.1}\% & \underline{4.66} & \underline{0.601} & \underline{37.8}\% & \textbf{12.48} \\
 & Comb. VPRAG+BRAG (Ours) & \textbf{0.346} & \textbf{0.537} & \textbf{0.643} & 0.913 & \textbf{0.508} & \textbf{50.4}\% & \textbf{4.93} & \textbf{0.606} & \textbf{45.9}\% & \underline{12.42} \\
\hline
\end{tabular}
}
\end{table*}
\begin{table*}[h]
\centering
\caption{Method Comparison Across LLMs (Garden Editing Prompt)}
\label{tab:method_comparison_stacked_garden}
\resizebox{\textwidth}{!}{
\begin{tabular}{l|l|cccc|ccc|ccc}
\hline
\multirow{3}{*}{\textbf{LLM}} & \multirow{3}{*}{\textbf{Method}} & \multicolumn{4}{c|}{\textbf{Individual Metrics}} & \multicolumn{3}{c|}{\textbf{$\mathrm{VPTT_{score}\text{-}c}$} (Uniform Weights)} & \multicolumn{3}{c}{\textbf{$\mathrm{VPTT_{score}}$} (Novelty Adjusted)} \\
& & \textbf{PA} & \textbf{GS} & \textbf{CP} & \textbf{NV} & \textbf{Score} & \textbf{Win\%} & \textbf{d} & \textbf{Score} & \textbf{Win\%} & \textbf{d} \\
& & & & & & & & \textit{(vs base)} & & & \textit{(vs base)} \\
\hline
\multirow{5}{*}{GPT-4o-mini} & Baseline & 0.131 & 0.364 & 0.588 & -- & 0.361 & 0.0\% & -- & 0.312 & 0.0\% & -- \\
 & Persona Only & 0.358 & 0.407 & 0.622 & -- & 0.462 & 0.2\% & 3.59 & 0.380 & 0.0\% & 2.98 \\
 & BRAG & 0.311 & \textbf{0.594} & \underline{0.650} & \underline{0.867} & 0.518 & 15.4\% & 4.33 & 0.609 & 17.0\% & 10.57 \\
 & VPRAG (Ours) & \textbf{0.403} & 0.582 & 0.650 & \textbf{0.901} & \textbf{0.545} & \textbf{46.2}\% & \textbf{5.12} & \textbf{0.630} & \textbf{52.6}\% & \textbf{11.47} \\
 & Comb. VPRAG+BRAG (Ours) & \underline{0.378} & \underline{0.588} & \textbf{0.663} & 0.860 & \underline{0.543} & \underline{38.2}\% & \underline{4.83} & \underline{0.623} & \underline{30.5}\% & \underline{10.92} \\
\hline
\multirow{5}{*}{Qwen} & Baseline & 0.131 & 0.364 & 0.588 & -- & 0.361 & 0.0\% & -- & 0.312 & 0.0\% & -- \\
 & Persona Only & 0.349 & 0.407 & 0.618 & -- & 0.458 & 0.2\% & 2.93 & 0.377 & 0.0\% & 2.52 \\
 & BRAG & 0.377 & \textbf{0.625} & \underline{0.668} & \underline{0.549} & \underline{0.557} & \underline{27.4}\% & \underline{4.90} & 0.573 & 12.0\% & \underline{6.85} \\
 & VPRAG (Ours)& \underline{0.403} & 0.583 & 0.653 & \textbf{0.858} & 0.546 & 20.9\% & \textbf{5.08} & \textbf{0.623} & \textbf{68.6}\% & \textbf{11.05} \\
 & Comb. VPRAG+BRAG (Ours) & \textbf{0.419} & \underline{0.613} & \textbf{0.678} & 0.529 & \textbf{0.570} & \textbf{51.6}\% & 4.50 & \underline{0.577} & \underline{19.3}\% & 6.39 \\
\hline
\multirow{5}{*}{Gemini} & Baseline & 0.131 & 0.364 & 0.588 & -- & 0.361 & 0.0\% & -- & 0.312 & 0.0\% & -- \\
 & Persona Only & 0.313 & 0.403 & 0.615 & -- & 0.444 & 0.6\% & 2.97 & 0.368 & 0.0\% & 2.49 \\
 & BRAG & 0.251 & \underline{0.563} & 0.639 & \underline{0.850} & 0.484 & 14.8\% & 2.88 & 0.581 & 14.8\% & 8.52 \\
 & VPRAG (Ours) & \textbf{0.340} & 0.544 & \underline{0.651} & \textbf{0.913} & \underline{0.511} & \underline{31.0}\% & \underline{3.94} & \textbf{0.609} & \textbf{46.2}\% & \textbf{10.24} \\
 & Comb. VPRAG+BRAG (Ours) & \underline{0.336} & \textbf{0.582} & \textbf{0.667} & 0.822 & \textbf{0.528} & \textbf{53.6}\% & \textbf{4.04} & \underline{0.606} & \underline{39.0}\% & \underline{9.66} \\
\hline
\end{tabular}
}
\end{table*}

{
    \small
    \bibliographystyle{ieeenat_fullname}
    \bibliography{main}
}


\end{document}